\documentclass[sigconf]{acmart}
\AtBeginDocument{%
  }

\copyrightyear{2025}
\acmYear{2025}
\setcopyright{cc}
\setcctype{by}
\acmConference[MM '25]{Proceedings of the 33rd ACM International Conference on Multimedia}{October 27--31, 2025}{Dublin, Ireland}
\acmBooktitle{Proceedings of the 33rd ACM International Conference on Multimedia (MM '25), October 27--31, 2025, Dublin, Ireland}\acmDOI{10.1145/3746027.3755607}
\acmISBN{979-8-4007-2035-2/2025/10}

\newcommand{\eg}{\textit{e}.\textit{g}. }

\usepackage{bbding}
\usepackage{subcaption}
\usepackage{multirow}
\usepackage{makecell} 
\usepackage{colortbl}
\usepackage{algorithm}
\usepackage{algorithmic}
\usepackage{threeparttable}
\usepackage{hyperref}
\usepackage{etoolbox}
\usepackage{balance}
\usepackage{cuted}

\newcommand{\tableref}[1]{%
  \ifstrequal{#1}{tab:proj}{Table \hyperref[tab:proj]{6}}%
  {\ifstrequal{#1}{tab:construct}{Table \hyperref[tab:construct]{5}}%
  {\ifstrequal{#1}{tab:ddc-learning}{Table \hyperref[tab:ddc-learning]{4}}%
  {\ifstrequal{#1}{tab:ablatecomponent}{Table \hyperref[tab:ablatecomponent]{3}}%
  {\tablename~\ref{#1}}}}}%
}

\begin{document}

\title{Advancing Reliable Test-Time Adaptation of Vision-Language Models under Visual Variations}

\author{Yiwen Liang}
\affiliation{%
  \institution{Tsinghua University}
  \city{Beijing}
  \country{China}
  }
\email{evenliang789@gmail.com}
\orcid{0009-0004-8936-4632}

\author{Hui Chen}
\authornote{Corresponding author.}
\affiliation{%
  \institution{Tsinghua University}
  \city{Beijing}
  \country{China}}
\email{jichenhui2012@gmail.com}
\orcid{0000-0003-4180-5801}

\author{Yizhe Xiong}
\affiliation{%
  \institution{Tsinghua University}
  \city{Beijing}
  \country{China}
 }
\email{xiongyizhe2001@163.com}
\orcid{0009-0001-5233-9466}

\author{Zihan Zhou}
\affiliation{%
  \institution{Tsinghua University}
  \city{Beijing}
  \country{China}
}
\email{blurry.kokan@gmail.com}
\orcid{0009-0007-3658-1002}

\author{Mengyao Lyu}
\affiliation{%
  \institution{Tsinghua University}
  \city{Beijing}
  \country{China}
  }
\email{mengyao.lyu@outlook.com}
\orcid{0000-0002-5404-4127}

\author{Zijia Lin}
\affiliation{%
  \institution{Tsinghua University}
  \city{Beijing}
  \country{China}
  }
\email{linzijia07@tsinghua.org.cn}
\orcid{0000-0002-1390-7424}

\author{Shuaicheng Niu}
\affiliation{%
  \institution{Nanyang Technological University}
  \city{Singapore}
  \country{Singapore}
  }
\email{shuaicheng.niu@ntu.edu.sg}
\orcid{0000-0001-8212-1831}

\author{Sicheng Zhao}
\affiliation{%
  \institution{Tsinghua University}
  \city{Beijing}
  \country{China}
  }
\email{schzhao@gmail.com}
\orcid{0000-0001-5843-6411}

\author{Jungong Han}
\affiliation{%
  \institution{Tsinghua University}
  \city{Beijing}
  \country{China}
  }
\email{jungonghan77@gmail.com}
\orcid{0000-0003-4361-956X}

\author{Guiguang Ding}
\authornotemark[1]             
\affiliation{%
  \institution{Tsinghua University}
  \city{Beijing}
  \country{China}}
\email{dinggg@tsinghua.edu.cn}
\orcid{0000-0003-0137-9975}

\renewcommand{\shortauthors}{Yiwen Liang et al.}

\begin{abstract}
    Vision-language models (VLMs) exhibit remarkable zero-shot capabilities but struggle with distribution shifts in downstream tasks when labeled data is unavailable, which has motivated the development of Test-Time Adaptation (TTA) to improve VLMs' performance during inference without annotations.
    Among various TTA approaches, cache-based methods show promise by preserving historical knowledge from low-entropy samples in a dynamic cache and fostering efficient adaptation.
    However, these methods face two critical reliability challenges: 
    (1) entropy often becomes \textit{unreliable} under distribution shifts, causing error accumulation in the cache and degradation in adaptation performance;
    (2) the final predictions may be \textit{unreliable} due to inflexible decision boundaries that fail to accommodate large downstream shifts.
    To address these challenges, we propose a Reliable Test-time Adaptation (ReTA)\footnote{Code is available at https://github.com/Evelyn1ywliang/ReTA.} method that integrates two complementary strategies to enhance reliability from two perspectives.  
    First, to mitigate the unreliability of entropy as a sample selection criterion for cache construction, we introduce Consistency-aware Entropy Reweighting (CER), which incorporates consistency constraints to weight entropy during cache updating. 
    While conventional approaches rely solely on low entropy for cache prioritization and risk introducing noise, our method leverages predictive consistency to maintain a high-quality cache and facilitate more robust adaptation. 
    Second, we present Diversity-driven Distribution Calibration (DDC), which models class-wise text embeddings as multivariate Gaussian distributions, enabling adaptive decision boundaries for more accurate predictions across visually diverse content.
    Extensive experiments demonstrate that ReTA consistently outperforms state-of-the-art methods, particularly under real-world distribution shifts.
\end{abstract}


\maketitle

\section{Introduction}
Large-scale vision-language models (VLMs), such as CLIP~\cite{clip} and ALIGN~\cite{align}, pre-trained on massive web-scale datasets, have demonstrated impressive zero-shot capabilities and strong open-world visual understanding across a wide range of vision tasks, including classification~\cite{scpnet,HSPNet,lsnet}, retrieval~\cite{raclip_retrieval,reco_retrieval}, and segmentation~\cite{rpn_seg,consolidator}. 
Their core mechanism involves learning a shared representation space by aligning visual and textual modalities during training, and subsequently employing similarity-based matching for classification at inference. 
However, VLMs often suffer from performance degradation when deployed on unlabeled test data from downstream domains that exhibit significant distribution shifts from their pre-training distribution~\cite{vlms1,learnfrom,pyra}, limiting their effectiveness in real-world scenarios with diverse visual variations.

Test-time adaptation (TTA) strategies have recently emerged to boost VLMs' capacity for adaptation to downstream out-of-distribution scenarios without relying on labeled data. 
Initial efforts focused on prompt-based approaches~\cite{tpt,difftpt,promptalign}, which adapt vision-language models by learning domain-specific prompts to minimize the entropy of predictions from augmented test samples.
While effective, these methods typically require iterative backpropagation through the entire encoder, leading to substantial computational overhead.
Recently, cache-based TTA methods~\cite{tda,boostadapter,dmn,dpe} have gained significant attention for their efficiency and are rapidly emerging as a dominant paradigm.
These approaches typically construct a dynamic cache that is updated online as new data arrives. 
High-confidence samples are selected based on the entropy of VLM predictions to populate the cache, which serves as historical knowledge to calibrate model predictions either in a training-free manner or with minimal parameter adjustments.
For example, TDA~\cite{tda} introduces dynamic caches to generate both positive and negative predictions for precise prediction refinement. 
DMN\cite{dmn} constructs complementary memory banks to exploit both long-term and short-term knowledge for enhanced adaptation. 
DPE~\cite{dpe} employs dual-modal residual learning to extract more accurate multi-modal representations while maintaining alignment between visual and textual prototypes.
These cache-based approaches offer notable computational efficiency with minimal overhead, making them particularly well-suited for resource-constrained applications.

\begin{figure}[!t]
    \centering
    \begin{subfigure}{0.23\textwidth}
        \centering
        \includegraphics[width=3.6cm]{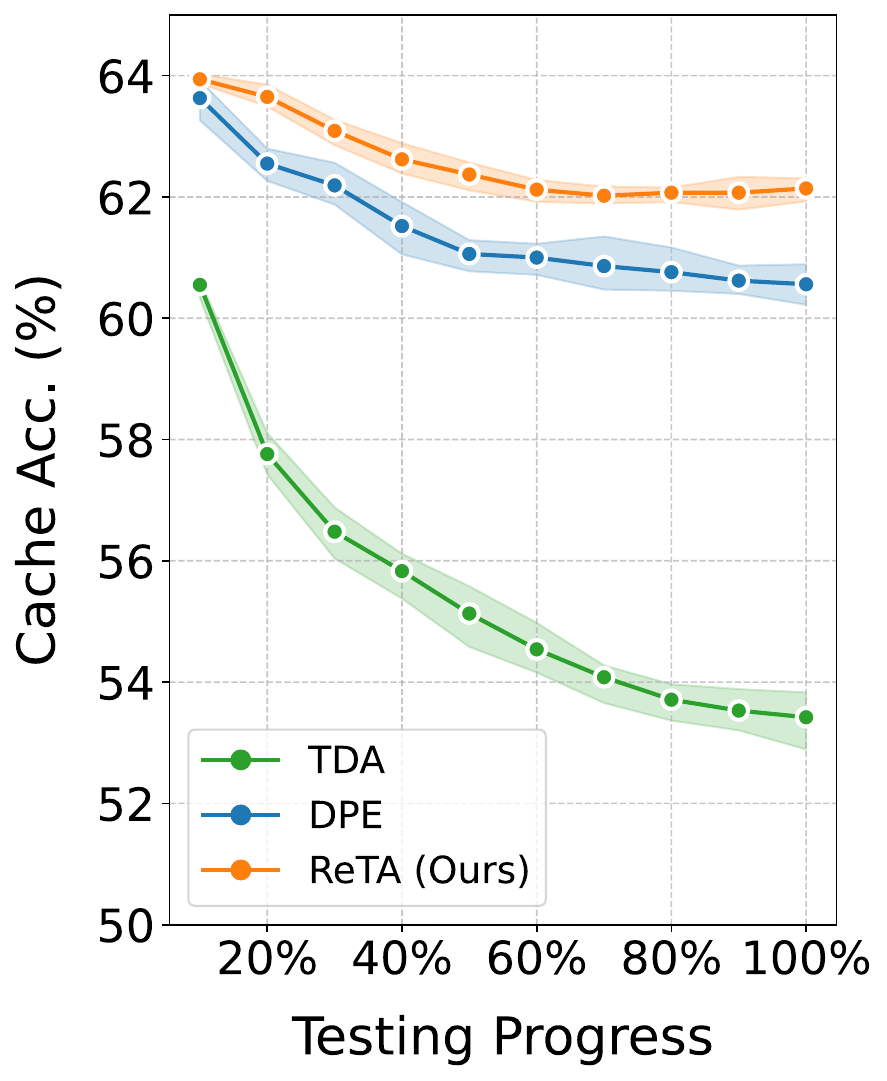}
        \caption{\hspace*{-2em}}
        \label{fig:1a}
    \end{subfigure}
    \hfill
    \begin{subfigure}{0.23\textwidth}
        \centering
        \includegraphics[width=3.6cm]{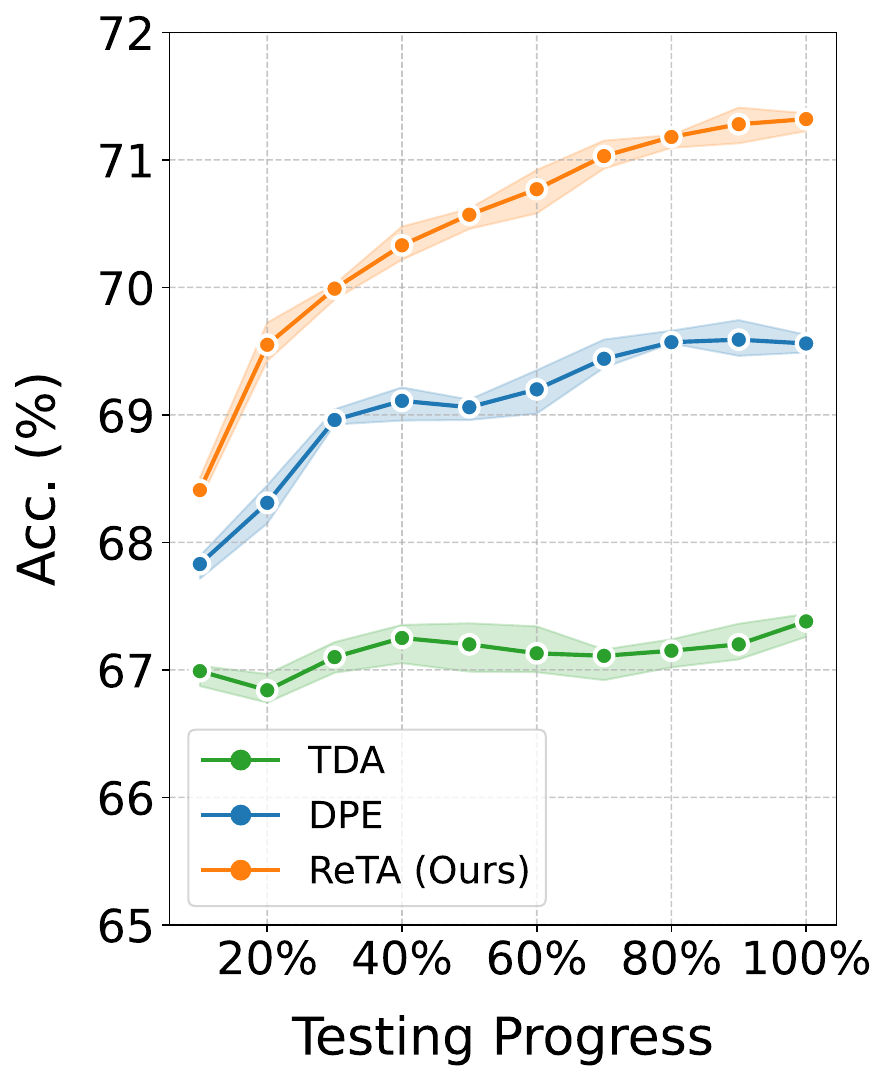}
        \caption{\hspace*{-2em}}
        \label{fig:1b}
    \end{subfigure}
    \vspace{-0.4cm}
    \caption{Impact of cache reliability on test-time adaptation performance. (a): Accuracy of samples stored in the cache during testing. (b): Accuracy of test samples during testing. As adaptation progresses, baseline methods show significant degradation in cache quality, while our proposed ReTA consistently maintains more reliable samples and further improves prediction accuracy under distribution shifts via adaptive decision boundaries.}
    \label{fig:1}
    \vspace{-0.5cm}
\end{figure}

\begin{figure}[!t]
\centering
\centerline{\includegraphics[width=7.3cm]{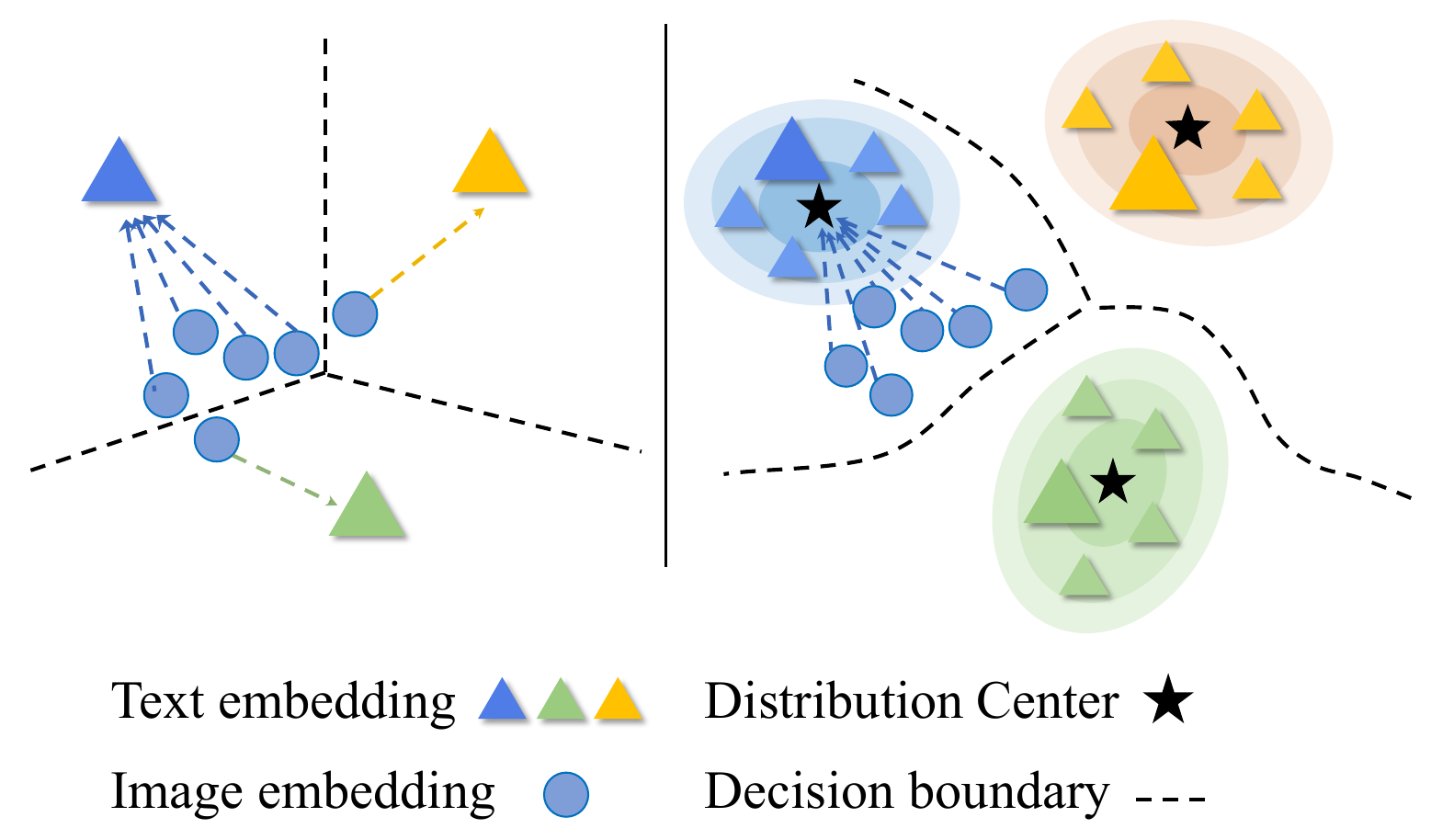}} 
    \vspace{-0.3cm}
\caption{Decision boundary visualization. 
Left: Conventional cache-based TTA uses fixed text embeddings, creating rigid boundaries that struggle with distribution shifts.
Right: ReTA implements adaptive distribution modeling for more flexible boundaries that better accommodate visual variations.}
    \label{fig: 2}
    \vspace{-0.5cm}
\end{figure}

Despite advances in cache-based TTA methods, reliable adaptation of VLMs under significant distribution shifts---particularly visual variations such as lighting changes, style differences, or domain gaps between training and testing---remains a critical challenge.
First, these approaches~\cite{boostadapter,tda,dpe} generally rely on entropy for cache prioritization. However, prior studies~\cite{entropy_unreliable,upl,zero} show that entropy-based measures become unreliable in unlabeled test-time scenarios, leading to suboptimal cache construction.
Furthermore, our observations indicate that the accuracy of samples stored in the cache progressively declines as testing proceeds (as shown in Figure~\ref{fig:1a}), further validating the unreliability of entropy as a measure of sample quality.
These low-quality samples inevitably introduce noise into the cache, causing the cached class-wise representations to diverge from their true distribution over time, which significantly limits performance improvements (see Figure~\ref{fig:1b}).
Second, the decision boundaries of VLMs become unreliable under distribution shifts to unlabeled target domains.
Even with TTA, these models struggle to adapt effectively, as their boundaries lack the flexibility to accommodate such shifts, leading to unreliable predictions.
Although some methods~\cite{dmn,dpe} attempt to refine the decision boundaries using mean textual prototypes derived from enriched prompts, they still lack the flexibility to handle intra-class diversity shifts, exemplified by the misalignment of blue image embeddings from their corresponding blue text embeddings in Figure~\ref{fig: 2}.

To address these issues, we propose Reliable Test-time Adaptation (ReTA), a method designed to improve the reliability of VLMs under distribution shifts caused by natural visual variations. 
ReTA integrates an entropy reweighting strategy and a distribution calibration technique to enhance adaptation performance.
Specifically, we introduce Consistency-aware Entropy Reweighting (CER) to identify reliable samples for cache prioritization.
CER constructs adjacent class-specific textual representations as a voting committee~\cite{boxlevel,commit1,commit2}, evaluating prediction consistency across multiple semantic perspectives to identify reliable samples.
By prioritizing semantically consistent samples during cache updates, CER promotes the retention of high-quality image features in the cache, mitigating distribution drift and enhancing adaptation stability.
Furthermore, to improve the reliability of predictions under visual diversity, we propose Diversity-driven Distribution Calibration (DDC).
DDC leverages adjacent text embeddings constructed by CER to extend each class representation from a single prototype to an approximate multivariate Gaussian distribution.
Through residual learning and progressive updates, DDC enables more flexible decision boundaries that better accommodate visual diversity and continuously adapt to feature variations during test-time adaptation.

The contributions of this work are summarized as follows:
\begin{itemize}
    \item We propose ReTA, a unified test-time adaptation method designed to enhance the reliability of vision-language models under significant distribution shifts in test-time adaptation.
    \item We introduce Consistency-aware Entropy Reweighting (CER), which constructs class-specific textual representations to evaluate prediction consistency across adjacent representations. CER identifies high-consistency samples and prioritizes them during cache updates to enhance adaptation.
    \item We develop Diversity-driven Distribution Calibration (DDC), which models text embeddings as approximate multivariate Gaussian distributions with residual learning and progressive updates, enabling adaptive decision boundaries that support reliable predictions under visual variations.
    \item Our comprehensive evaluation across multiple benchmarks demonstrates that ReTA consistently outperforms state-of-the-art TTA methods and is effective in adapting VLMs.
    
\end{itemize}

\section{Related Works}

\subsection{Vision-Language Models}
Large-scale pre-trained vision-language models (VLMs), such as CLIP~\cite{clip}, have demonstrated remarkable capabilities in learning generalizable visual and textual representations. 
Numerous efforts have focused on adapting these powerful pre-trained models to downstream tasks.
Early approaches like CoOp~\cite{coop} and CoCoOp~\cite{cocoop} achieved notable gains through lightweight prompt learning with limited supervision in few-shot scenarios. 
More recent works, such as Tip-Adapter~\cite{tipadapter} and TaskRes~\cite{taskres}, introduced efficient adaptation strategies by leveraging feature memory constructed from a small set of labeled samples.
These memory-based methods have attracted considerable attention recently for avoiding full model fine-tuning while enabling rapid domain adaptation with minimal overhead. 
Despite their strong performance and low data requirements, they still rely on labeled target-domain data, which limits their applicability in real-world scenarios where annotations are unavailable.
In contrast, our work focuses on test-time adaptation, where models must generalize to novel domains without any labeled data during inference, better reflecting real-world constraints and advancing truly adaptive vision-language systems.

\subsection{Vision-Language Test-Time Adaptation}
Test-time adaptation (TTA) enhances the adaptability of vision-language models (VLMs) to target domains using only unlabeled test samples~\cite{tpt,swapprompt,rlcf,zero}.
Recent approaches primarily fall into two categories: prompt-based and cache-based methods.
Prompt-based methods optimize continuous textual embeddings during inference. TPT~\cite{tpt} fine-tunes prompts by enforcing consistency across augmented views of the same image, while DiffTPT~\cite{difftpt} introduces diffusion-based augmentations to increase view diversity and improve model adaptability to unseen test data.
Cache-based approaches leverage historical instance-level information from test samples.
TDA~\cite{tda} constructs positive and negative feature caches from test samples to refine CLIP predictions via combined similarity scores. 
DMN~\cite{dmn} integrates static and dynamic memory networks to exploit both few-shot and historical unlabeled data, using a flexible interactive strategy for improved adaptation. 
BoostAdapter~\cite{boostadapter} incorporates regional-level bootstrapping for more precise memory construction.
DPE~\cite{dpe} jointly evolves visual and textual prototypes to better capture domain-specific semantics. 
While prompt-based methods offer simplicity, cache-based strategies generally achieve stronger adaptation and maintain high computational efficiency by more effectively modeling target domain distributions through lightweight test-time memory. 
Our work builds on the cache-based paradigm, with advancements in cache quality via reliable sample selection and improving prediction robustness through adaptive decision boundary calibration in dynamic environments.


\begin{figure*}[htb]
	
	\begin{minipage}[b]{1.0\linewidth}
		\centering
        \vspace{-0.18cm}
		\centerline{\includegraphics[width=13.6cm]{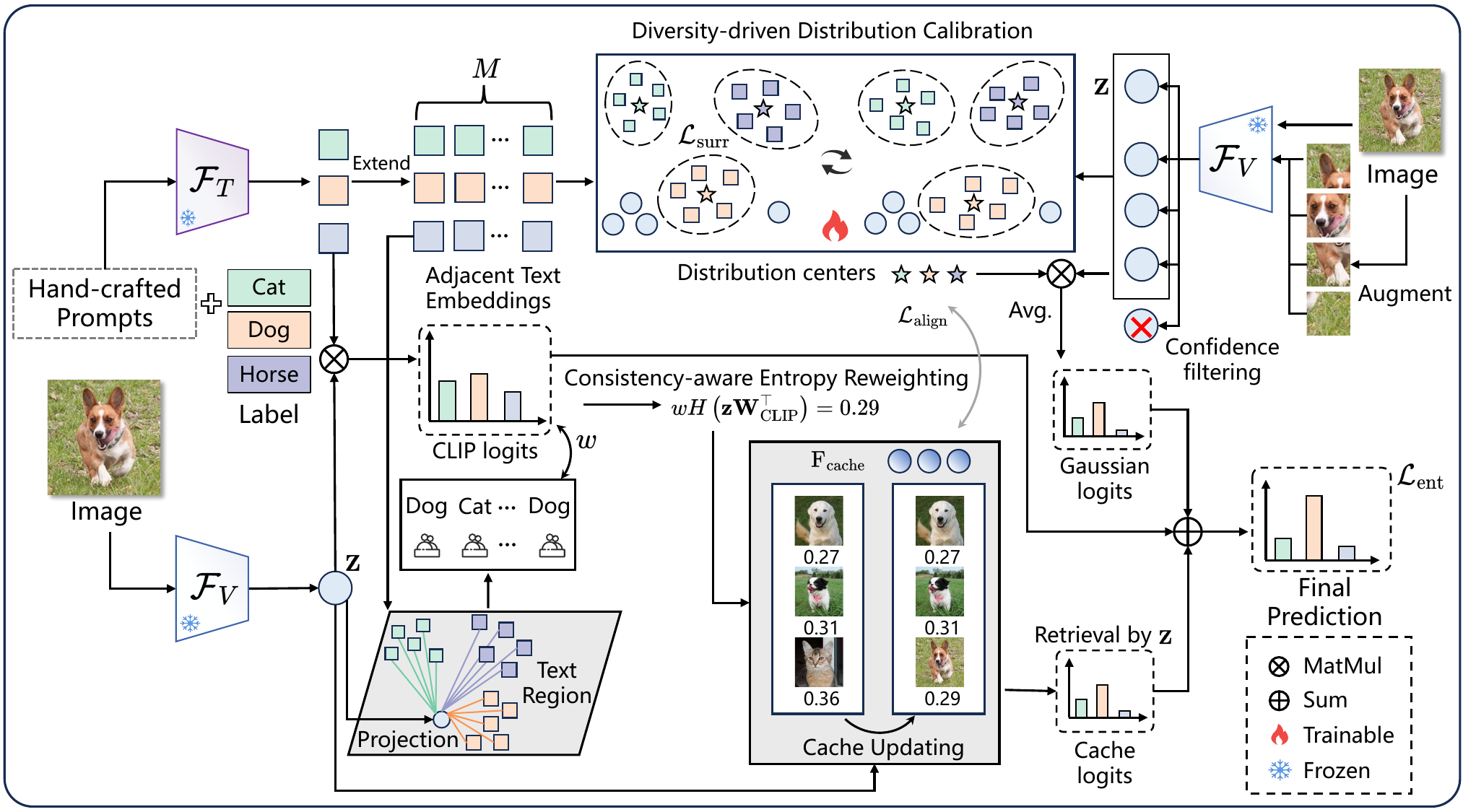}}
	\end{minipage}
        \vspace{-0.7cm}
	\caption{The overall framework of our proposed Reliable Test-time Adaptation (ReTA) approach under visual variations. 
    ReTA leverages multiple text prompts to construct adjacent embeddings and adjust entropy-based prioritization via consistent predictions, thereby preserving more reliable samples in the cache. 
    For subsequent robust decision-making, ReTA models multivariate Gaussian distributions that are dynamically updated using test images and their augmentations, creating flexible boundaries that better accommodate changing visual content during test time.}
	\label{fig: 3}
	\vspace{-0.35cm}
\end{figure*}

\section{Methodology}

The pipeline of our proposed method is illustrated in Figure~\ref{fig: 3}. Sec.~\ref{3.1} presents preliminaries on CLIP and cache-based test-time prediction. 
Then, we propose Consistency-aware Entropy Reweighting (CER) in Sec.~\ref{3.2}, which selects reliable samples based on prediction consistency to construct a high-quality cache.
To enhance adaptation to diverse visual content, we introduce Diversity-driven Distribution Calibration (DDC) in Sec.~\ref{3.3}, which models class representations as approximate Gaussian distributions. 
Finally, Sec.~\ref{3.4} details our unified learning process and inference algorithm, showing how they train and inference in an end-to-end manner.

\subsection{Preliminaries} \label{3.1}

\textbf{Zero-shot CLIP Prediction.} CLIP is pre-trained on large-scale image-text pairs to align visual and textual modalities within a unified embedding space by maximizing cosine similarity through a contrastive loss. 
CLIP consists of a visual encoder $\mathcal{F}_V$ and a text encoder $\mathcal{F}_T$, which project inputs into a shared latent space $\mathbb{R}^d$, where $d$ is the feature dimension, for cross-modal alignment.
During inference, an image $x$ is classified by computing similarity between the image feature $\mathbf{z} = \mathcal{F}_V(x)$, and textual embeddings $\boldsymbol{t}^c=\mathcal{F}_T(\mathcal{P}^c)$, where $\mathcal{P}^c$ denotes a text prompt describing class $c$, and $C$ is the total number of classes.
The zero-shot CLIP prediction probability is then given by:
\begin{equation}
    \setlength{\abovedisplayskip}{1.5pt}
    \setlength{\belowdisplayskip}{2.0pt}
    \label{eq:clip}
    p^c_{\scriptsize \text{CLIP}}=\frac{\exp \left(\mathbf{z} \cdot \boldsymbol{t}^c / \tau\right)}{\sum_{j=1}^{C} \exp \left(\mathbf{z} \cdot \boldsymbol{t}^j / \tau\right)} 
\end{equation}
where $\tau$ is the temperature parameter, typically set to 0.01.

\textbf{Cache-based Test-time Prediction. }Recent approaches leverage cache mechanisms at test time for efficient vision-language model adaptation. 
Building upon the key-value storage paradigm of TIP-Adapter~\cite{tipadapter}, many cache-based TTA methods~\cite{tda,dpe,boostadapter} have emerged, dynamically storing and updating high-confidence feature-label pairs during inference.
The cache model functions as a priority queue, ranked by entropy derived from CLIP prediction.
With $\mathbf{W}_{\mathrm{CLIP}}=\left[\boldsymbol{t}^1, \boldsymbol{t}^2, \ldots, \boldsymbol{t}^C\right]^{\top}$ as stacked text embeddings serving as the classifier, each cache item is represented as:
\begin{equation}
    \left\{\mathbf{F}_{\text {cache }}, \mathbf{L}_p, H\left(\mathbf{F}_{\text {cache }} \mathbf{W}_{\scriptsize \text {CLIP }}^{\top}\right)\right\},
\end{equation}
where $\mathbf{F}_{\text {cache }}$ is the stored image feature, $\mathbf{L}_p$ is a one-hot pseudo-label obtained from CLIP prediction, and the entropy $H(p)$ is:
\begin{equation}
    \scalebox{0.96}{$
    H(p) = \left(-\sum_{c=1}^{C} p^c_{\scriptsize \text{CLIP}} \log p^c_{\scriptsize \text{CLIP}}\right).
    $}
\end{equation}
For streaming test-time samples arriving on the fly, new items enter the corresponding cache slots based on their pseudo-labels. Once the cache reaches capacity, items with the highest entropy are replaced.
At inference, a new test feature $\mathbf{z}$ retrieves relevant cache items,  and the final prediction combines CLIP logits with cache-based logits:
\begin{equation}
\label{eq:cls_logits}
\text{logits}_{\mathrm{cls}}(\mathbf{z}) = \mathbf{z}\mathbf{W}_\text{CLIP}^{\top} + \mathcal{A}(\mathbf{z}\mathbf{F}_{\text{cache}}^{\top})\mathbf{L}_p
\end{equation}
where $\mathcal{A}(x)=\alpha \exp (-\beta(1-x))$ is a modulating function with scaling factor $\alpha$ and sharpness ratio $\beta$.
Following DPE~\cite{dpe}, we compute $\mathbf{F}_{\text{cache}}$ as the mean of all samples within each cache slot, forming class-specific visual prototypes that become more compact and representative as more samples are accumulated.

\subsection{Consistency-aware Entropy Reweighting} \label{3.2}

Previous methods~\cite{dpe,tda,boostadapter} take low-entropy cached samples as visual prototypes for feature retrieval and cache prediction computation. 
However, relying solely on entropy is unreliable as CLIP is usually overconfident, potentially assigning high confidence to misclassified samples~\cite{zero,entropy_unreliable}. Since prediction confidence lacks correlation with pseudo-label accuracy~\cite{upl,cdisvit}, low-entropy samples may carry incorrect labels, resulting in cached prototypes deviating from the expected distribution and degrading retrieval performance.
While approaches like TDA~\cite{tda} and DPE~\cite{dpe} have incorporated knowledge-enriched text embeddings with different prompts, they typically average them into a single representation, which fails to fully exploit their semantic diversity.

To address this limitation, we propose Consistency-aware Entropy Reweighting (CER), which improves cache quality by prioritizing the storage of relatively reliable samples. 
Specifically, CER leverages prediction consistency across neighboring semantic representations to identify reliable and representative samples.
Inspired by the committee-based strategies~\cite{boxlevel,commit1,commit2}, we construct a semantic voting committee using multiple text embeddings to assess prediction reliability.
While controversial samples expose uncertainty in committee members' judgments, CER prioritizes those with consistent agreement in voting, as they are more representative and reliable for caching.

\textbf{Adjacent Class-specific Textual Embedding Construction.} 
To measure the predictive consistency of samples, we first construct a set of adjacent class-specific textual embeddings. Following~\cite{tda,dpe,dmn}, we consider multiple class-specific prompts for each class $c$ to build text representation sets. 
Formally, we define the prompt set for class $c$ as $\boldsymbol{\mathcal{P}}^c=\left\{\mathcal{P}_1^c, \mathcal{P}_2^c, \ldots, \mathcal{P}_K^c\right\}$, where $\mathcal{P}_i^c$ is the $i$-th prompt with corresponding text embedding $\boldsymbol{t}_i^c=\mathcal{F}_T\left(\mathcal{P}_i^c\right)$ from CLIP's text encoder.
Rather than simply averaging prompts, we leverage subtle similarity differences among them to construct semantic embedding sets that form local neighborhoods in text space, enabling assessment of prediction consistency via representational neighborhood consistency~\cite{neighboor,neighbour2}. 
The relative similarity of each prompt to others within the same class is computed as:
\begin{equation}
    \setlength{\abovedisplayskip}{1.0pt}
    \setlength{\belowdisplayskip}{1.0pt}
    sim_i^c = \sum_{j=1, j \neq i}^{K} \cos(\boldsymbol{t}_i^c, \boldsymbol{t}_j^c)
\end{equation}
where $sim_i^c$ represents the cumulative intra-class cosine similarity of the $i$-th text embedding.

We then sort text embeddings by relative cosine similarity in ascending order, arranging them from semantic outliers to centroids, and construct our final adjacent text embedding set $\left\{\widehat{\boldsymbol{t}}_1^c, \widehat{\boldsymbol{t}}_2^c, \ldots, \widehat{\boldsymbol{t}}_M^c\right\}$ with $M$ embeddings per class using a progressive binning approach. 
Specifically, we define $Q_m = \lfloor \frac{mK}{M} \rfloor$ for $m = 1,2,...,M$ to determine embeddings for the $m$-th adjacent embedding. Each $\widehat{\boldsymbol{t}}_m^c$ is created by averaging the first $Q_m$ embeddings from the sorted sequence, where embeddings are cumulatively included in progressively expanding pools. 
Our ascending progressive binning strategy creates a smooth semantic transition across embeddings while preserving diversity, thus forming a natural semantic neighborhood for subsequent consistency assessment.
Functionally, these adjacent text embeddings operate as a semantic voting committee, where each embedding offers a diverse perspective on class prediction.

\textbf{Projection for Modality Consistency.} 
To effectively leverage textual neighborhood consistency for identifying reliable  image features, we project image features into the text embedding space to mitigate the modality gap~\cite{modalitygap,modalitygap2}. 
Given the adjacent text embeddings $\tilde{\mathcal{T}}=\left\{\widehat{\boldsymbol{t}}_m^c: 1 \leq c \leq C, 1 \leq m \leq M\right\}$, we decompose this collection via Singular Value Decomposition (SVD) to extract principal semantic variation directions:
\begin{equation}
\setlength{\abovedisplayskip}{1.5pt}
\setlength{\belowdisplayskip}{1.0pt}
U \Sigma V^T = \text{SVD}(\tilde{\mathcal{T}})
\end{equation}
where $U$, $\Sigma$, and $V$ represent the left singular vectors, the singular values (in descending order), and the right singular vectors, respectively.
By retaining the top-$n$ principal components, we collect primary singular vectors $\tilde{V} \in \mathbb{R}^{n \times d}$ to capture the text feature space's intrinsic semantic structure, forming the text subspace with projection matrix:
\begin{equation}
\setlength{\abovedisplayskip}{1.5pt}
\setlength{\belowdisplayskip}{2.0pt}
\Phi_\text{proj} = \tilde{V}^T \tilde{V} \in \mathbb{R}^{d \times d}.
\end{equation}
For structural coherence and compact cross-modal representations, we project each image feature into the text subspace as~\cite{proj2}:
\begin{equation}
{\mathbf{z}}_\text{proj}=\Phi_\text{proj} \mathbf{z}
\end{equation}

\textbf{Prediction Consistency Assessment.} We assess prediction consistency by examining classification variations across semantic neighborhoods. For each test data, we create a set of pseudo-labels by computing similarity between the projected image feature and each adjacent text embedding:
\begin{equation} \label{eq:10}
\hat{\mathcal{Y}}= \left\{ \hat{y}_1, \hat{y}_2, ..., \hat{y}_M \right\}, \text{where }  \hat{y}_m = \arg\max\nolimits_{c \in {C}} \, ({\mathbf{z}}_\text{proj}^T \widehat{\boldsymbol{t}}_m^c)
\end{equation}
where $\hat{y}_m$ is the pseudo-label for the $m$-th adjacent text embedding. 
We derive a \textbf{stability-consistency score} incorporating both prediction stability and consistency with the original prediction. 
We first determine the most frequent prediction within $\hat{\mathcal{Y}}$ as an indicator of stability:
\begin{equation}
    \setlength{\abovedisplayskip}{1.0pt}
    \setlength{\belowdisplayskip}{1.2pt}
y^\ast = \arg\max\nolimits_{c \in {C}}\sum_{m=1}^{M} \mathbb{I}(\hat{y}_m = c)
\end{equation}
where $y^\ast$ represents the majority-voted pseudo-label and $\mathbb{I}(\cdot)$ is the indicator function. 
Next, we assess consistency between this voted pseudo-label and the original prediction with a consistency factor:
\begin{equation}
    \setlength{\abovedisplayskip}{1.0pt}
    \setlength{\belowdisplayskip}{1.5pt}
    \text{Consistency: }\mathcal{R} = 
    \begin{cases}
    1, & \text{if } y^\ast = y \\
    \gamma, & \text{if } y^\ast \neq y
    \end{cases}
\end{equation}
where $y = \arg\max\nolimits_{c \in {C}} (\mathbf{z}^T \widehat{\boldsymbol{t}}_M^c)$ denotes the original prediction based on $\mathbf{z}$ and the final text prototype, and $\gamma > 1$ is a penalty parameter that increases entropy, thereby reducing the priority of inconsistent predictions.
Additionally, we define a stability factor $\mathcal{S}$:
\begin{equation}
    \setlength{\abovedisplayskip}{0.8pt}
    \setlength{\belowdisplayskip}{1.5pt}
    \text{Stability: }\mathcal{S} = \frac{M}{n^\ast}, \quad 1 \leq \mathcal{S} \leq M
\end{equation}
where $n^\ast$ is the count of the most common prediction. 
A larger $\mathcal{S}$ indicates less stable predictions, with $\mathcal{S} = 1$ denoting perfect agreement. 
The stability-consistency score is calculated as:
\begin{equation} \label{eq:14}
w = 1+\log(\mathcal{R} \mathcal{S})
\end{equation}

\textbf{Entropy Reweighting for Cache Updating.} Finally, we compute the reweighted entropy using our stability-consistency score:
\begin{equation}
\setlength{\abovedisplayskip}{2.0pt}
\setlength{\belowdisplayskip}{2.0pt}
H'(\mathbf{z}) = w \cdot H(\mathbf{z})
\end{equation}
where $H(\mathbf{z})$ is the entropy and $p = \text{Softmax}({\mathbf{z}}^{\top} \boldsymbol{\widehat{t}}_M^c)$. 
The defined score $w$ amplifies the entropy of samples with unstable or inconsistent predictions, thereby penalizing unreliable samples.
This reweighted entropy replaces the standard entropy in the cache priority queue, enabling more reliable cache updates:
\begin{equation}
    \left\{\mathbf{F}_{\text {cache }}, \mathbf{L}_p, H'(\mathbf{F}_{\text {cache }}\mathbf{W}_{\text {CLIP }}^{\top})\right\}.
\end{equation}
CER prioritizes features with both low entropy and high consistency. When the cache reaches capacity, items with the highest values of reweighted entropy are replaced first, filtering out unreliable samples and preserving higher-quality class prototypes for robust adaptation under visual variations.

 \begin{table*}[!t]
  \centering
  \caption{Performance evaluation on cross-dataset generalization. Results report top-1 accuracy (\%) across different datasets for both CLIP-RN50 and CLIP-ViT-B/16.}
    \vspace{-0.4cm}
    \begin{tabular}{lcccccccccc|c}
    \toprule
    Method & Caltech & DTD   & Cars  & EuroSAT & Aircraft & Flowers & Pets  & UCF101 & Food101 & SUN397 & Average \\
    \midrule
    CLIP-RN50 & 85.88 & 40.37 & 55.70 & 23.69 & 15.66 & 61.75 & 83.57 & 58.84 & 73.97 & 58.80 & 55.82 \\
    Ensemble & 87.26 & 40.37 & 55.89 & 25.79 & 16.11 & 62.77 & 82.97 & 59.48 & 74.82 & 60.85 & 56.63 \\
    CoOp~\cite{coop}  & 86.53 & 37.29 & 55.32 & 26.20 & 15.12 & 61.55 & 87.00 & 59.05 & 75.59 & 58.15 & 56.18 \\
    TPT~\cite{tpt}    & 87.02 & 40.84 & 58.46 & 28.33 & 17.58 & 62.69 & 84.49 & 60.82 & 74.88 & 61.46 & 57.66 \\
    DiffTPT~\cite{difftpt}  & 86.89 & 40.72 & 60.71 & 41.04 & 17.60 & 63.53 & 83.40 & 62.67 & \textbf{79.21} & 62.72 & 59.85 \\
    TDA~\cite{tda}   & 89.70 & 43.74 & 57.78 & 42.11 & 17.61 & 68.74 & 86.18 & 64.18 & 77.75 & 62.53 & 61.03 \\
    BoostAdapter~\cite{boostadapter} & 88.48 & 43.85 & 59.67 & \textbf{44.40} & 18.93 & 68.25 & 85.75 & 64.42 & 78.78 & 62.83 & 61.54 \\
    DPE~\cite{dpe}   & \textbf{90.83} & 50.18 & 59.26 & 41.67 & 19.80 & 67.60 & 85.97 & 61.98 & 77.83 & 64.23 & 61.93 \\
    \rowcolor{gray!20}
    ReTA  & 90.35 & \textbf{52.46} & \textbf{61.11} & 39.64 & \textbf{22.62} & \textbf{70.12} & \textbf{86.90} & \textbf{66.18} & 77.46 & \textbf{65.11} & \textbf{63.20} \\
    \midrule
    CLIP-ViT-B/16 & 93.35 & 44.27 & 65.48 & 42.01 & 23.67 & 67.44 & 88.25 & 65.13 & 83.65 & 62.59 & 63.58 \\
    Ensemble & 93.55 & 45.04 & 66.11 & 50.42 & 23.22 & 66.99 & 86.92 & 65.16 & 82.86 & 65.63 & 64.59 \\
    CoOp~\cite{coop}  & 93.70 & 41.92 & 64.51 & 46.39 & 18.47 & 68.71 & 89.14 & 66.55 & 85.30 & 64.15 & 63.88 \\
    TPT~\cite{tpt}   & 94.16 & 47.75 & 66.87 & 42.44 & 24.78 & 68.98 & 87.79 & 68.04 & 84.67 & 65.50 & 65.10 \\
    DiffTPT~\cite{difftpt} & 92.49 & 47.00 & 67.01 & 43.13 & 25.60 & 70.10 & 88.22 & 62.67 & 87.23 & 65.74 & 65.47 \\
    TDA~\cite{tda}   & 94.24 & 47.40 & 67.28 & 58.00 & 23.91 & 71.42 & 88.63 & 70.66 & 86.14 & 67.62 & 67.53 \\
    Zero-Ensemble~\cite{zero} & 94.14 & 45.86 & 68.48 & 42.09 & 24.42 & 66.82 & 87.20 & 68.57 & 84.58 & 66.90 & 64.91 \\
    BoostAdapter~\cite{boostadapter} & 94.77 & 45.69 & \textbf{69.30} & \textbf{61.22} & 27.45 & 71.66 & 89.51 & 71.93 & 87.17 & 68.09 & 68.68 \\
    DPE~\cite{dpe}   & 94.81 & 54.20 & 67.31 & 55.79 & 28.95 & 75.07 & 91.14 & 70.44 & 86.17 & 70.07 & 69.40 \\
    \rowcolor{gray!20}
    ReTA  & \textbf{95.29} & \textbf{57.39} & 69.11 & 58.26 & \textbf{31.86} & \textbf{77.55} & \textbf{92.37} & \textbf{74.52} & \textbf{86.69} & \textbf{70.70} & \textbf{71.37} \\
    \bottomrule
    \vspace{-0.5cm}
    \end{tabular}%
  \label{tab:cross}%
\end{table*}%


\subsection{Diversity-driven Distribution Calibration} \label{3.3}
While CER provides reliable features for cache-based adaptation, it remains challenging to make rapid adjustments to each test-time input, as the cache contains only a few high-confidence samples and may be limited in accommodating visual variations.
Existing cache-based methods~\cite{tda,boostadapter,dmn} attempt to compensate for this drawback by enhancing discriminability through sophisticated prompts, but their adaptation capabilities remain limited in dynamic environments due to VLMs' fixed decision boundaries, which struggle to handle distribution shifts. 
To overcome this challenge, we propose Diversity-driven Distribution Calibration (DDC), which leverages adjacent text embeddings from CER to construct a dynamically adjustable distribution. 
Drawing from recent success in modeling prompt embeddings with Gaussian distributions~\cite{proda,prodalike}, we characterize the adjacent text embeddings $\tilde{\mathcal{T}}$ as an approximate Gaussian distribution $\mathcal{N}\left(\boldsymbol{\mu}^c, \boldsymbol{\Sigma}^c\right)$, where $\boldsymbol{\mu}^c$ representing the mean embedding of class $c$ and $\boldsymbol{\Sigma}^c$ denoting the corresponding covariance matrix. 
Building on the efficient residual learning strategy in DPE~\cite{dpe}, we extend this idea to adapt flexible representational distributions, enabling DDC to accommodate evolving visual features.

\textbf{Textual Gaussian Distribution Evolution via Residuals.} 
For efficient adaptation while preserving well-defined semantic boundaries, we employ residual learning for text embeddings. Specifically, we evolve the class-wise Gaussian distribution by adjusting each adjacent embedding through residual updates. Residual parameters are introduced as:
\begin{equation}
    \setlength{\abovedisplayskip}{4.0pt}
    \setlength{\belowdisplayskip}{2.0pt}
    \label{eq:residual}
    \widehat{\boldsymbol{t}}_m^c = \frac{\widehat{\boldsymbol{t}}_m^c + \boldsymbol{r}_m^c}{||\widehat{\boldsymbol{t}}_m^c + \boldsymbol{r}_m^c||}
\end{equation}
where $\boldsymbol{r}_m^c$ represents the learnable residual with zero initialization. 
Unlike DPE, which refines both textual and visual prototypes at the instance level, our DDC performs distribution-level optimization on textual embeddings, offering more adaptive decision boundaries. 
To continuously adapt to distribution shifts in downstream domains, we maintain a counter $l$ for tracking confident samples and update the textual representation following the progressive strategy in~\cite{dpe}:
\begin{equation}
    \setlength{\abovedisplayskip}{2.0pt}
    \setlength{\belowdisplayskip}{2.0pt}
    \label{updating}
    \widehat{\boldsymbol{t}}_m^c = \frac{(l-1)\widehat{\boldsymbol{t}}_m^c + \widehat{\boldsymbol{t}}_{m}^{c(\ast)}}{||(l-1)\widehat{\boldsymbol{t}}_m^c + \widehat{\boldsymbol{t}}_{m}^{c(\ast)}||}
\end{equation}
where $\widehat{\boldsymbol{t}}_{m}^{c(\ast)}$ represents the newly optimized embedding from Eq.~\ref{eq:residual}.
To ensure reliable adaptation, we perform local residual updating (Eq.~\ref{eq:residual}) only for samples deemed reliable by CER (those with $w=1$). 
For global updates (Eq.~\ref{updating}), we apply a fixed threshold $\tau_c$ to discard updates from low-confidence samples (for which $H(\mathbf{z}) < \tau_c$), ensuring stability in the evolving global representation.

\textbf{Objective for Distribution Calibration. }To achieve distribution calibration, we jointly optimize textual embeddings through a comprehensive objective that integrates instance-level residual learning and distribution-level supervision:
\begin{equation} \label{eq:allloss}
    \setlength{\abovedisplayskip}{2.5pt}
    \setlength{\belowdisplayskip}{2.5pt}
\mathcal{L} = \mathcal{L}_{\text{ent}} + \lambda_1 \mathcal{L}_{\text{surr}} + \lambda_2 \mathcal{L}_{\text{align}}
\end{equation}
where $\lambda_1, \lambda_2$ are weighting parameters. 
The entropy $\mathcal{L}_{\text{ent}}$ encourages confident predictions from a test sample and its $N$ augmentations with respect to each text embedding, defined as:
\begin{equation}
    \setlength{\abovedisplayskip}{1.5pt}
    \setlength{\belowdisplayskip}{2.0pt}
    \scalebox{0.90}{$
    \begin{aligned}
    & \mathcal{L}_{\text{ent}} = H(\tilde{p}_{\text{cls}}(\mathbf{x})) = -\sum_{c=1}^C \tilde{p}_{\text{cls}}^c \log(\tilde{p}_{\text{cls}}^c) \\
    & \text{with} \; \tilde{p}^{\text{cls}}_c = \frac{\sum_{i=1}^N \mathbb{I}[H(p_{\text{cls}}^c(\mathbf{x}_i)) < \delta] \cdot p_{\text{cls}}^c(\mathbf{x}_i)}{\sum_{i=1}^N \mathbb{I}[H(p_{\text{cls}}^c(\mathbf{x}_i)) < \delta]}
    \end{aligned}
    $}
\end{equation}
where $p_{\text{cls}}^c$ is the probability derived from Eq.~\ref{eq:cls_logits} and $\delta$ is a entropy threshold. 
To perform reliable distribution-level calibration, we select pseudo-labels $\tilde{y}$ corresponding to reliable samples with $w = 1$ (Eq.\ref{eq:14}) for learning, and leverage multiple augmented views to capture variations. We adopt a surrogate loss following ProDA~\cite{proda}:
\begin{equation}
    \setlength{\abovedisplayskip}{2.0pt}
    \setlength{\belowdisplayskip}{2.0pt}
    \scalebox{0.82}{$
    \begin{aligned}
    \mathcal{L}_{\text{surr}} & = \underset{(\mathbf{z}, \tilde{y})}{\mathbb{E}}\left[-\log \frac{\exp \left(\left\langle\mathbf{z}, \widehat{\boldsymbol{t}}_M^{\tilde{y}}\right\rangle / \tau\right)}{\sum_{c=1}^C \exp \left(\left\langle\mathbf{z}, \widehat{\boldsymbol{t}}_M^c\right\rangle / \tau+\mathbf{z}^T \mathbf{W}_{c, \tilde{y}} \mathbf{z} / 2 \tau^2\right)}\right]
    \end{aligned}
    $}
\end{equation}
where covariance term $\small\mathbf{W}_{c, \tilde{y}}=\boldsymbol{\Sigma}^{c c}+\boldsymbol{\Sigma}^{\tilde{y} \tilde{y}}-\boldsymbol{\Sigma}^{c \tilde{y}}-\boldsymbol{\Sigma}^{\tilde{y} c}$ models class-wise variations. 
To encourage alignment between textual and visual distributions, we further introduce a cross-modal alignment loss:
\begin{equation}
\setlength{\abovedisplayskip}{2.5pt}
\scalebox{0.92}{$
    \mathcal{L}_{\text{align}} = \frac{1}{C} \sum_{c=1}^C \left(-\log \frac{\exp((\widehat{\boldsymbol{t}}_M^c)^{\top} \mathbf{F}_{\text{cache}}^c)}{\sum_{j=1}^C \exp((\widehat{\boldsymbol{t}}_M^c)^{\top} \mathbf{F}_{\text{cache}}^j)} -\log \frac{\exp((\widehat{\boldsymbol{t}}_M^c)^{\top} \mathbf{F}_{\text{cache}}^c)}{\sum_{j=1}^C \exp((\widehat{\boldsymbol{t}}_M^j)^{\top} \mathbf{F}_{\text{cache}}^c)}\right)
$}
\end{equation}
where $\small\mathbf{F}_{\text{cache}}^c$ is the average visual prototype of class $c$ from the cache. 
By jointly optimizing these objectives, our method calibrates class distributions to construct adaptive decision boundaries that robustly accommodate visual diversity.

\subsection{Unified Learning Process and Inference} \label{3.4}

The complete ReTA test-time adaptation process is detailed in the Algorithm 1 in Appendix. For each test sample, we first apply CER to assess prediction consistency and reweight its entropy, determining its cache storage prioritization. 
Then, for reliable samples, DDC cumulatively updates the residuals of the text embeddings throughout the entire adaptation process.

During inference, we follow ProDA~\cite{proda} by using the Gaussian mean as a refined decision boundary for classification:
\begin{equation}
    \setlength{\abovedisplayskip}{2.0pt}
    \setlength{\belowdisplayskip}{2.0pt}
    p_{\text{gauss}}=\frac{\exp \left(\left\langle\mathbf{z}, \boldsymbol{\mu}^c\right\rangle / \tau\right)}{\sum_{c=1}^C \exp \left(\left\langle\mathbf{z}, \boldsymbol{\mu}^c\right\rangle / \tau\right)}
    \end{equation}
where $\small\boldsymbol{\mu}^c=\frac{1}{M} \sum_{m=1}^M \widehat{\boldsymbol{t}}_m^c$ denotes the class-wise mean of the evolved text embeddings, and $p_{\text{gauss}}$ represents the refined probability derived from our Gaussian modeling.
For final prediction, ReTA integrates the outputs from both cache-based prediction in Eq.~\ref{eq:cls_logits} and our Gaussian modeling refined logits:
\begin{equation} \label{finalpred}
\setlength{\abovedisplayskip}{1.8pt}
\setlength{\belowdisplayskip}{2.0pt}
p_\text{final} = p_{\text{cls}} + \eta p_{\text{gauss}}
\end{equation}
where $p_{\text{cls}} = \text{softmax}(\text{logits}_{\mathrm{cls}}(\mathbf{z}))$ and $\eta$ balances the contributions. 
In summary, by unifying CER and DDC, ReTA effectively preserves high-quality samples and dynamically refines classifier boundaries, enabling more reliable and efficient test-time adaptation.


\begin{table}[!t]
  \centering
    \caption{Performance evaluation on robustness to natural distribution shifts. Results report top-1 accuracy (\%) for both CLIP-RN50 and CLIP-ViT-B/16.}
    \vspace{-0.4cm}
    \resizebox{\columnwidth}{!}{
    \begin{threeparttable}
    \begin{tabular}{lccccccc}
    \toprule
    Method & Imagenet & -A    & -V2   & -R    & -Sketch & Avg.  & OOD Avg. \\
    \midrule
    CLIP-RN50 & 58.16 & 21.83 & 51.41 & 56.15 & 33.37 & 44.18 & 40.69 \\
    Ensemble & 59.81 & 23.24 & 52.91 & 60.72 & 35.48 & 46.43 & 43.09 \\
    CoOp~\cite{coop}  & 63.33 & 23.06 & 55.40 & 56.60 & 34.67 & 46.61 & 42.43 \\
    TPT~\cite{tpt}   & 60.74 & 26.67 & 54.70 & 59.11 & 35.09 & 47.26 & 43.89 \\
    DiffTPT~\cite{difftpt} & 60.80 & 31.06 & 55.80 & 58.80 & 37.10 & 48.71 & 45.69 \\
    TDA~\cite{tda}   & 61.35 & 30.29 & 55.54 & 62.58 & 38.12 & 49.58 & 46.63 \\
    TPS~\cite{tps}   & 61.47 & 30.48 & 54.96 & 62.87 & 37.14 & 49.38 & 46.36 \\
    DMN-ZS~\cite{dmn} & 63.87 & 28.57 & 56.12 & 61.44 & 39.84 & 49.97 & 46.49 \\
    BoostAdapter\tnote{1}~\cite{boostadapter} & 61.04 & \textbf{35.52} & 56.22 & 62.87 & 38.87 & 50.91 & 48.37 \\
    DPE~\cite{dpe}  & 63.41 & 30.15 & \textbf{56.72}  & 63.72 & 40.03 & 50.81 & 47.66 \\
    \rowcolor{gray!20}
    ReTA  & \textbf{63.90} & 34.42 & 56.34 & \textbf{64.10} & \textbf{40.54} & \textbf{51.86} & \textbf{48.85} \\
    \midrule
    CLIP-ViT-B/16 & 66.73 & 47.87 & 60.86 & 73.98 & 46.09 & 59.11 & 57.20 \\
    Ensemble & 68.34 & 49.89 & 61.88 & 77.65 & 48.24 & 61.20 & 59.42 \\
    CoOp~\cite{coop}  & 71.51 & 49.71 & 64.20 & 75.21 & 47.99 & 61.72 & 59.28 \\
    TPT~\cite{tpt}   & 68.98 & 54.77 & 63.45 & 77.06 & 47.94 & 62.44 & 60.81 \\
    DiffTPT~\cite{difftpt} & 70.30 & 55.68 & 65.10 & 75.00 & 46.80 & 62.28 & 60.52 \\
    TDA~\cite{tda}   & 69.51 & 60.11 & 64.67 & 80.24 & 50.54 & 65.01 & 63.89 \\
    TPS~\cite{tps}   & 70.19 & 60.08 & 64.73 & 80.27 & 49.95 & 65.04 & 63.76 \\
    Zero-Ensemble~\cite{zero} & 70.93 & 64.06 & 65.16 & 80.75 & 50.32 & 66.24 & 65.07 \\
    DMN-ZS~\cite{dmn} & \textbf{72.25} & 58.28 & 65.17 & 78.55 & 53.20 & 65.49 & 63.80 \\
    BoostAdapter\tnote{1}~\cite{boostadapter} & 69.39 & 64.06 & 65.19 & 80.77 & 51.51 & 66.19 & 65.39 \\
    DPE~\cite{dpe}   & 71.91 & 59.63 & 65.44 & 80.40 & 52.26 & 65.93 & 64.43 \\
    \rowcolor{gray!20}
    ReTA  & 72.18 & \textbf{64.22} & \textbf{65.52} & \textbf{81.31} & \textbf{53.21} & \textbf{67.29} & \textbf{66.07} \\
    \bottomrule
    \end{tabular}%
        \begin{tablenotes}
        \footnotesize
        \item[1] We reproduce the BoostAdapter result because the original paper does not provide ImageNet results.  
      \end{tablenotes}

    \end{threeparttable}
    }
  \label{tab:ood}%
  \vspace{-0.3cm}
\end{table}%

\section{Experiments}

\subsection{Experimental Setup}

\textbf{Datasets.} We conduct experiments on two widely used benchmarks, following prior works~\cite{tpt,promptalign,tda}: Cross-Datasets Generalization and Robustness to Natural Distribution Shifts. 
The first benchmark comprises 10 diverse image classification datasets: Aircraft~\cite{fgvc}, Caltech101~\cite{caltech}, Cars~\cite{cars}, Describable Textures (DTD)~\cite{dtd}, EuroSAT~\cite{eurosat}, Flowers102~\cite{flowers}, Food101~\cite{food101}, Pets~\cite{pets}, SUN397~\cite{sun397}, and UCF101~\cite{ucf101}. 
The second benchmark consists of the ImageNet~\cite{imagenet} and its four variants: ImageNetV2~\cite{imagenetv}, ImageNet-R~\cite{imagenetr}, ImageNet-Sketch~\cite{imagenets}, and ImageNet-A~\cite{imageneta}.
Following the setup of TPT~\cite{tpt}, we report the top-1 accuracy on each dataset.

\textbf{Baselines}. In our experiments, we compare the performance of the proposed ReTA with several state-of-the-art TTA methods, including: (1) TPT~\cite{tpt}; (2) DiffTPT~\cite{difftpt}; (3) TDA~\cite{tda}; (4) DMN-ZS~\cite{dmn}; (5) TPS~\cite{tps}; (6) BoostAdapter~\cite{boostadapter}; (7) Zero~\cite{zero}; and (8) DPE~\cite{dpe}.
Results for the baselines are directly taken from their publications unless otherwise noted.

\textbf{Implementation details.} 
In our main experiments, we adopt the pre-trained CLIP-ViT-B/16 and CLIP-RN50 as the base models. 
The batch size is set to 1, with each test image augmented to 63 additional views through AugMix~\cite{augmix}, with random resized cropping and random horizontal flipping as in TPT~\cite{tpt}, resulting in 64 images per batch.
The normalized entropy threshold is fixed at 0.1. We set the cache size to 3 samples per class, the number of adjacent text embeddings to $M=3$, and retain $n=64$ singular vectors from the SVD decomposition. The weighting factors $\lambda_1$ and $\lambda_2$ are set to 0.3 and 0.02 respectively.
We conduct all experiments with three different random seeds and report the average result. 
Unless otherwise specified, our analytical experiments report the average accuracy on Cross-Datasets with CLIP-ViT-B/16. 
Additional details are provided in the Appendix.

\subsection{Evaluation of Cross-Datasets Generalization}
As demonstrated in Table \ref{tab:cross}, we evaluate ReTA's transferability across diverse domains using the Cross-Datasets benchmark.
With the CLIP-RN50 backbone, ReTA achieves an average accuracy of 63.20\%, surpassing most existing methods and demonstrating competitive performance.
When using CLIP-ViT-B/16, ReTA consistently excels across the benchmark, achieving the best performance on 8 out of 10 tasksand attaining a notable 71.37\% average accuracy. 
In particular, ReTA shows substantial improvements on more challenging datasets, such as the texture-rich DTD (57.39\%) and fine-grained domains like Aircraft (31.86\%) and Flowers (77.55\%).
These compelling gains highlight ReTA's strong generalization ability in fine-grained and diverse cross-domain scenarios.

\subsection{Evaluation of Robustness to Natural Distribution Shifts}
We further evaluate ReTA on in-domain ImageNet and its four out-of-distribution variants to assess its robustness under natural distribution shifts. As shown in Table~\ref{tab:ood}, ReTA consistently outperforms current state-of-the-art methods, including BoostAdapter~\cite{boostadapter} and Zero-Ensemble~\cite{zero}, which serve as the two strongest baselines.
With CLIP-RN50, ReTA achieves an average accuracy of 51.86\%, surpassing the best-performing BoostAdapter (50.91\%) by 0.95\%.
When implemented with CLIP-ViT-B/16, ReTA demonstrates even more substantial gains, achieving the best performance across all OOD datasets. 
Specifically, it outperforms BoostAdapter by 1.10\% and Zero-Ensemble by 1.05\% on average across all five datasets, establishing a new state-of-the-art.  
These improvements validate the effectiveness of ReTA in achieving reliable adaptation under complex domain shift scenarios.

\begin{table}[!t]
    \captionsetup[subtable]{labelformat=empty}
    \centering
    \begin{subtable}{.22\textwidth}
        \centering
        \caption{Table 3: Ablations on ReTA components. ``NDS'': Natural Distribution Shifts; ``CD'': Cross-Datasets.}
        \vspace{-0.15cm}
        \resizebox{0.94\textwidth}{!}{
            \begin{tabular}{ll|c|c}
            \toprule
            \textbf{CER}   & \textbf{DDC}  & \textbf{NDS} & \textbf{CD} \\
            \midrule
            \XSolidBrush    & \XSolidBrush    & 66.04     & 69.79 \\
            \Checkmark  & \XSolidBrush    &  66.53     & 70.41 \\
            \XSolidBrush    & \Checkmark  & 66.92     & 70.91 \\
            \rowcolor{gray!20}
            \Checkmark  & \Checkmark  & \textbf{67.29} & \textbf{71.37} \\
            \bottomrule
            \end{tabular}%
            }
        \label{tab:ablatecomponent}%
    \end{subtable}%
    \hfill
    \begin{subtable}{.23\textwidth}
        \centering
        \caption{Table 4: Ablations on different learning mechanisms for DDC. TR: Training.}
        \vspace{-0.15cm}
        \resizebox{0.98\textwidth}{!}{
            \begin{tabular}{l|c}
            \toprule
            \textbf{Method} & \textbf{Acc. (\%)} \\
            \midrule
            w/o TR & 70.43 \\
            w/ TR, w/o $\mathcal{L}_{\text{ent}}$ & 70.72 \\
            w/ TR, w/ only $\mathcal{L}_{\text{ent}}$ & 71.01 \\
            \rowcolor{gray!20}
            w/ TR, w/ $\mathcal{L}_{\text{ent}}$ & \textbf{71.37} \\
            \bottomrule
            \end{tabular}
            }
        \label{tab:ddc-learning}
    \end{subtable}
    \label{tab:combined_ablation}
    \vspace{-0.2cm}
\end{table}

\begin{table}[!t]
    \captionsetup[subtable]{labelformat=empty}
    \centering
    \begin{subtable}{.26\textwidth}
        \centering
        \vspace{-0.15cm}
        \caption{Table 5: Comparison of different text embedding binning strategies.}
        \vspace{-0.15cm}
        \begin{tabular}{lc}
        \toprule
        \textbf{Binning Method} & \textbf{Acc. (\%)} \\
        \midrule
        \rowcolor{gray!20}
        Ascending progressive & \textbf{71.37} \\
        Descending progressive & 70.47 \\
        Uniform & 69.41 \\
        \bottomrule
        \end{tabular}
        \label{tab:construct}
    \end{subtable}%
    \hfill
    \begin{subtable}{.16\textwidth}
        \centering
        \vspace{-0.15cm}
        \caption{Table 6: Comparison of projection methods.}
        \vspace{-0.15cm}
        \begin{tabular}{lc}
        \toprule
        \textbf{Method} & \textbf{Acc. (\%)} \\
        \midrule
        \rowcolor{gray!20}
        SVD & \textbf{71.37} \\
        PCA & 70.57 \\
        LDA & 70.99 \\
        \bottomrule
        \end{tabular}
        \label{tab:proj}
    \end{subtable}
    \label{tab:combined}
    \vspace{-0.52cm}
\end{table}


\subsection{Experimental Analysis}
\textbf{Ablation on Main Components of ReTA. }We conduct a comprehensive analysis to examine the individual contributions of Consistency-aware Entropy Reweighting (CER) and Diversity-driven Distribution Calibration (DDC) on both two benchmarks.
As shown in \tableref{tab:ablatecomponent}, both components lead to clear performance improvements.
Notably, DDC has a greater impact, yielding gains of 0.88\% and 1.12\% on two benchmarks respectively. This highlights the effectiveness of DDC in refining decision boundaries, thereby enabling more reliable final predictions for test-time data with significant visual variations.
Additionally, integrating CER enhances the reliability of cache updating, fostering more accurate cache-based predictions. 
These two modules are complementary, with their combination yields synergistic improvements across both benchmarks.

\begin{figure}[!t]
	\centering
        \vspace{-0.1cm}
        \includegraphics[width=0.4\textwidth]{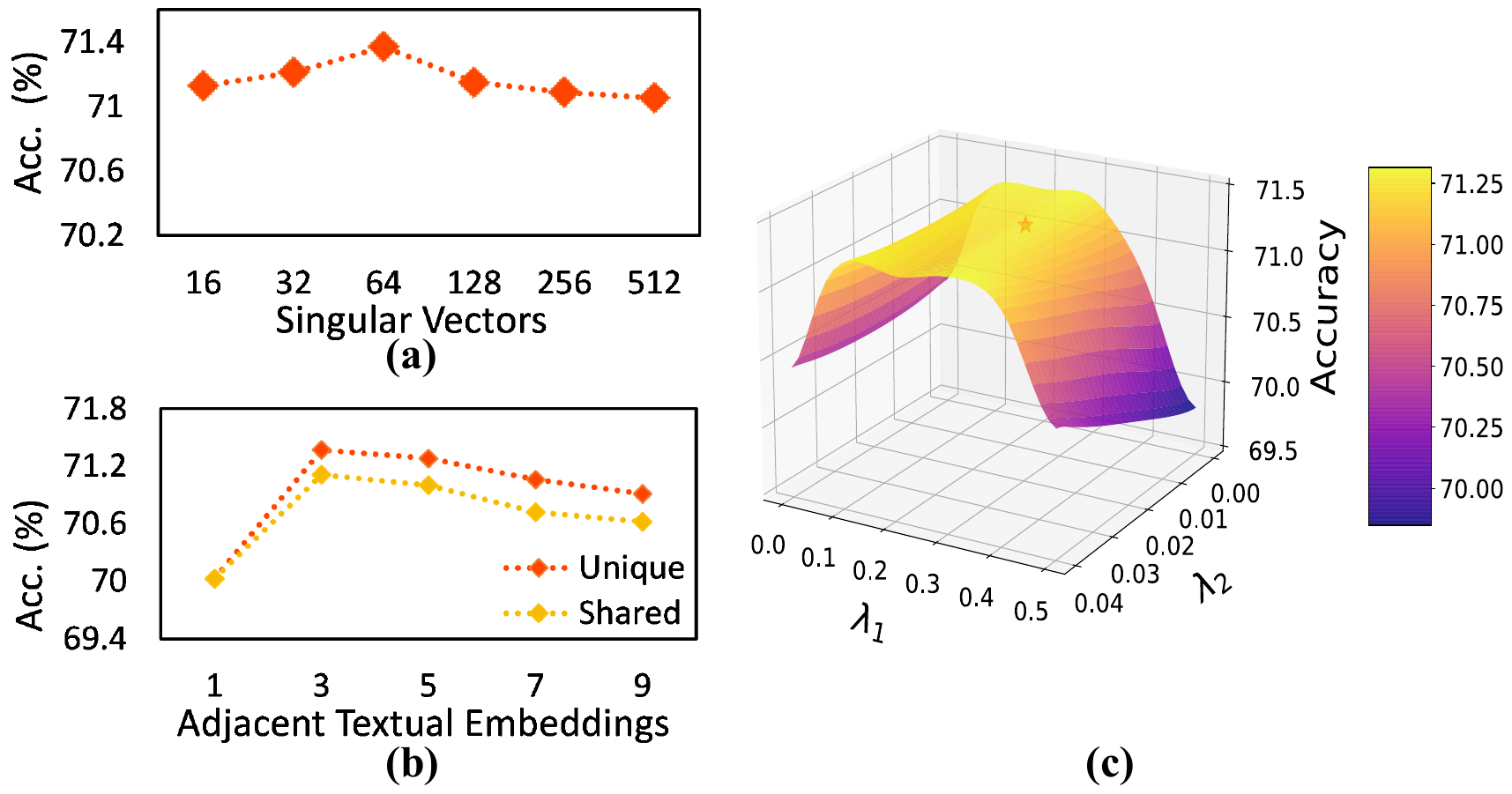}
        \vspace{-0.45cm}
        \caption{Hyperparameter analysis. (a) Number of singular vectors in SVD. (b) Number of adjacent textual embeddings. (c) Loss sensitivity analysis. }
        \label{fig:hyperparameter_ablation}
        \vspace{-0.4cm}
\end{figure}

\textbf{Analysis of Adjacent Textual Embeddings in CER. }
We analyze the adjacent textual embeddings introduced in Sec.~\ref{3.2} from two perspectives: (1) the effect of varying the number of adjacent embeddings ($M$), and (2) the impact of different construction methods.
For (1), the number of adjacent textual embeddings ($M$) determines both the size of the semantic voting committee in CER and the number of class-wise representations used in DDC for Gaussian modeling. 
As shown in Figure~\ref{fig:hyperparameter_ablation}b, using a single embedding ($M=1$) reduces the method to a standard cache-based TTA setting without adjacent structure, which performs worse due to its limited ability to model visual diversity and identify reliable samples.
Performance improves substantially as $M$ increases to 3, while larger values introduce redundancy without additional gains. 
Furthermore, assigning unique residuals to each embedding outperforms the shared-residual variant, as it enables more expressive modeling of intra-class variations.
For (2), \tableref{tab:construct} compares different construction methods for adjacent textual embeddings: (i) ascending progressive binning arranges text embeddings from semantic outliers to centroids based on relative cosine similarity; (ii) descending progressive binning follows the opposite order; (iii) uniform binning selects embeddings at equal intervals without considering semantic relationships. Our proposed ascending progressive binning achieves the best performance (71.37\%), outperforming descending and uniform binning. We attribute these improvements to the smoother semantic transitions formed by the ascending structure, which facilitates more coherent neighborhood construction and enables more reliable prediction consistency assessment.

\begin{figure}[!t]
	\centering
        \vspace{-0.12cm}
        \centerline{\includegraphics[width=6.8cm]{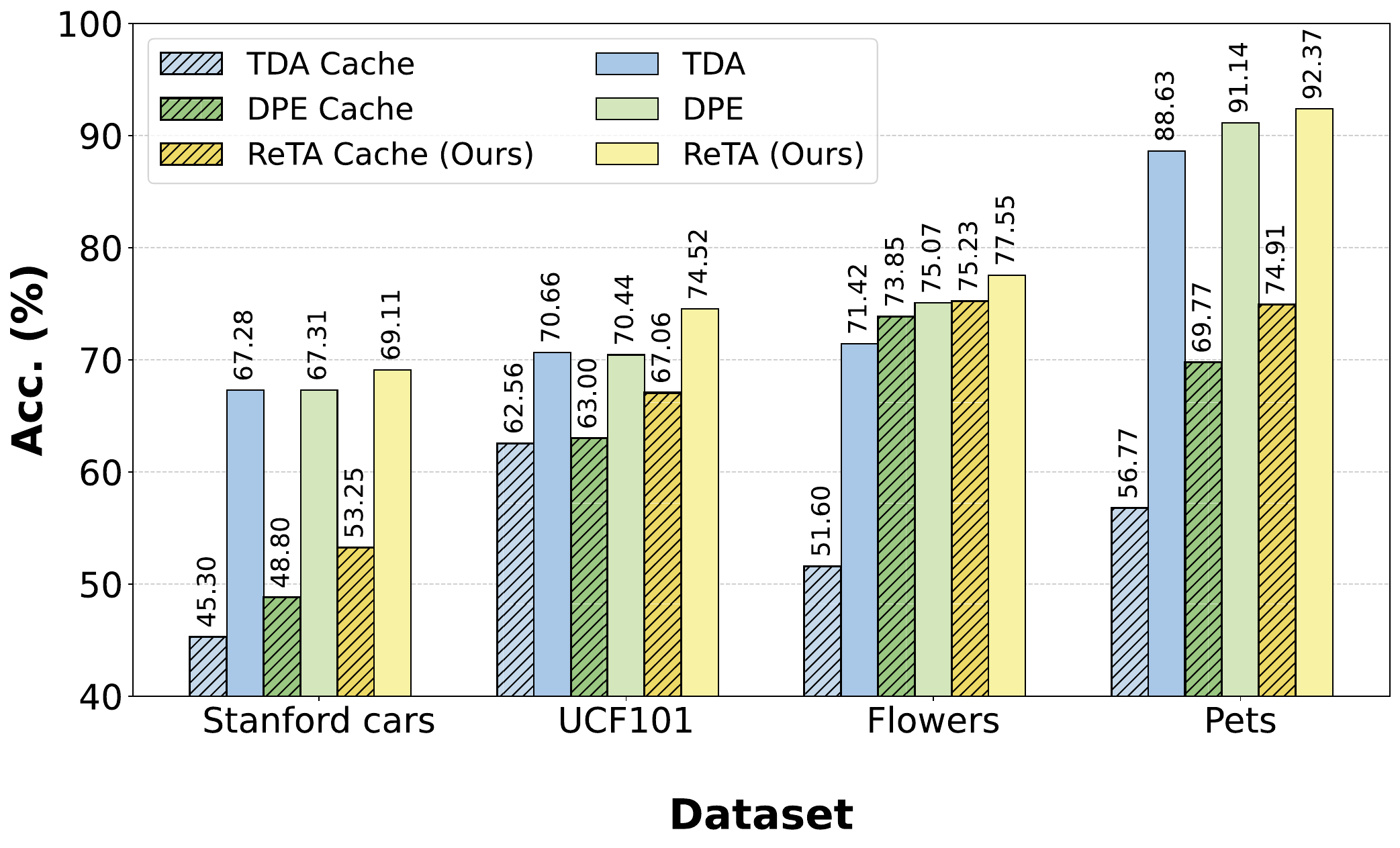}}
        \vspace{-0.45cm}
	\caption{Final comparison of adaptation performance and cache accuracy across four datasets for cache-based methods.}
        \label{fig: cachereliability}
        \vspace{-0.5cm}
\end{figure}

\textbf{Analysis of Projection in CER.} We analyze the effect of the number of singular vectors retained in the SVD projection, and compare different projection strategies. 
Figure~\ref{fig:hyperparameter_ablation}a shows an inverted U-shaped accuracy curve as the number of singular vectors increases. Retaining 64 components yields the best performance by balancing semantic relevance and noise reduction.
At this optimal dimension, \tableref{tab:proj} demonstrates that SVD projection outperforms PCA and LDA by 1.12\% and 0.39\%, respectively. 
SVD better preserves the underlying semantic structure without imposing strong class-dependent constraints. 
In contrast, PCA may retain excessive non-semantic variance, while LDA can be overly restrictive by focusing solely on discriminative class boundaries.

\textbf{Comparison of Cache Reliability and Final Performance.} Figure~\ref{fig: cachereliability} compares cache accuracy and overall classification performance across four fine-grained datasets. 
ReTA shows remarkable performance in both metrics, achieving 74.52\% on UCF101 (surpassing TDA by 3.86\% and DPE by 4.08\%) and 77.55\% on Flowers (exceeding TDA by 6.13\% and DPE by 2.48\%). 
Combined with evidence from Figure~\ref{fig:1}, these results prove that ReTA achieves superior final accuracy while maintaining high cache reliability.

\textbf{Effects of Learning Strategy in DDC.}
\tableref{tab:ddc-learning} investigates how different learning mechanisms within DDC affect performance. Without learning, DDC performs similarly to the CER-only baseline, indicating that static distribution modeling provides negligible benefits. 
Following prior methods~\cite{tpt,difftpt,dpe}, we adopt $\mathcal{L}_\text{ent}$ as the primary loss. It substantially improves performance by optimizing all textual residuals for more confident predictions.
While the auxiliary losses $\mathcal{L}_\text{surr}$ and $\mathcal{L}_\text{align}$ provide additional gains, properly balancing these components is crucial. 
As shown in Figure~\ref{fig:hyperparameter_ablation}c, the best performance is obtained at $\lambda_1{=}0.3$ and $\lambda_2{=}0.02$, where a higher $\lambda_1$ is critical for effective distribution-level calibration.

\setcounter{table}{6}
\begin{table}[!t] 
  \centering
  \caption{Testing Time and performance gains on robustness to natural distribution shifts.}
  \vspace{-0.42cm}
  \resizebox{\columnwidth}{!}{
      \begin{tabular}{l@{\hspace{5pt}} c@{\hspace{5pt}}c@{\hspace{5pt}}c@{\hspace{5pt}}c@{\hspace{5pt}}c@{\hspace{5pt}} c@{\hspace{5pt}} c}
        \toprule
        \multirow{2}{*}{\textbf{Method}} 
        & \multicolumn{5}{c}{\makecell{\textbf{Testing} \textbf{Time}}} 
        & \multirow{2}{*}{\textbf{Acc.}} 
        & \multirow{2}{*}{\textbf{Gain}} \\
        \cline{2-6}
          & \textbf{ImageNet} & \textbf{-A} & \textbf{-R} & \textbf{-V} & \textbf{-Sk.} 
          &  &  \\
        \hline
        CLIP         & 8min      & 1min      & 5min     &  2min     &  9min     & 59.11 & -    \\
        
        TPT~\cite{tpt}          & 10h   & 26min      &  1h50min     &  2h8min     &  11h5min     & 62.44 & 3.33 \\
                
        TDA~\cite{tda}          & 1h34min & 10min    & 52min      & 16min      & 1h41min      & 65.01 & 5.90 \\
        
        DPE~\cite{dpe}          & 3h41min & 34min    & 1h58min      & 44min      &  4h3min     & 65.93 & 6.82 \\
        
        BoostAdapter~\cite{boostadapter} & 3h25min & 13min    & 1h6min      &  51min     & 5h27min      & 66.34 & 7.23 \\
        \rowcolor{gray!20}
        ReTA         & 2h38min & 12min & 1h1min    & 29min      & 2h44min      & \textbf{67.29} & \textbf{8.18} \\
        \bottomrule
      \end{tabular}
  }
  \vspace{-0.28cm}
  \label{tab:time}
\end{table}

\textbf{Comparison of Computational Efficiency and Effectiveness.}
In Table~\ref{tab:time}, we evaluate the computational efficiency of ReTA and other test-time adaptation methods on ImageNet and its variants.
All methods are compared under AugMix augmentation aligned with TPT~\cite{tpt} for fair comparison, although this operation can be time-consuming. 
ReTA is 1.65x faster than BoostAdapter on large ImageNet and ImageNet-Sketch datasets, despite BoostAdapter being a training-free method like TDA. However, BoostAdapter introduces an additional loop for boosting-based cache updates, incurring significant overhead, especially on larger datasets. 
Compared to DPE, which also employs residual learning, ReTA is 1.56x faster while achieving better accuracy gains (+8.18 vs. +6.82) due to DPE's high computational cost from visual prototype evolution. 
By optimizing only textual residuals and eliminating the encoder backpropagation, ReTA achieves better optimization and a favorable balance between computational cost and performance gains.


\section{Conclusion}
In this paper, we introduce Reliable Test-time Adaptation (ReTA) to address two reliability issues in adapting VLMs: the unreliability of cached samples and inflexible decision boundaries. 
ReTA comprises two components: Consistency-aware Entropy Reweighting (CER), which improves cache quality via consistency-based sample selection, and Diversity-driven Distribution Calibration (DDC), which models text embeddings as Gaussians to construct adaptive decision boundaries for reliable predictions.
Experiments across multiple benchmarks show that ReTA consistently outperforms state-of-the-art methods, especially under challenging distribution shifts, 
offering a practical and robust solution in dynamic environments.


\begin{acks}
This work was supported by National Natural Science Foundation of China (Nos. 62525103, 62441235, 62271281, 62021002) and Beijing Natural Science Foundation (No. L252009).
\end{acks}

\bibliographystyle{ACM-Reference-Format}
\balance
\bibliography{ref}

\clearpage  
\nobalance

\appendix


\section*{\LARGE  Appendix}
\vspace{1.0em} 

\section{Full Algorithm of ReTA}

In Algorithm~\ref{alg:reta} we present the detailed procedures in ReTA.

\begin{algorithm}
\caption{ReTA: Reliable Test-time Adaptation}
\label{alg:reta}
    \begin{algorithmic}[1]
    \REQUIRE Test samples $\{\mathbf{x}_i\}$, CLIP encoders $\mathcal{F}_V$, $\mathcal{F}_T$, adjacent embeddings $\{\widehat{\boldsymbol{t}}_m^c\}_{m=1}^M$, cache size $SZ$, total classes $C$, projection matrix $\Phi_\text{proj}$, cached features $\mathbf{F}_{\text{cache}}$
    \STATE Initialize empty class-wise cache $\text{Cache}[c]$ for each class $c$ and set all residuals $\{\boldsymbol{r}_m^c \} = \mathbf{0}$
    \FOR{each test sample $\mathbf{x}_i$}
        \STATE $\mathbf{z}_i = \mathcal{F}_V(\mathbf{x}_i)$, $\mathbf{z}_{\text{proj}} = \Phi_\text{proj}\mathbf{z}_i$  
       \STATE Obtain pseudo label $y = \arg\max_{c \in {C}} (\mathbf{z}_i^{\small{\top}} \hat{\boldsymbol{t}}_M^c)$ 
        \STATE Calculate stability-consistency score $w$  \COMMENT{Eq. 13}
        \STATE Compute reweighted entropy $H'(\mathbf{z}_i) = w \cdot H(\mathbf{z}_i)$ 
        \IF{$|\text{Cache}[y]| < SZ$ OR $H'(\mathbf{z}_i) < \max \left( H'(\mathbf{F}_{\text{cache}}) \right)$}
            \STATE Update $\text{Cache}[y]$ with $(\mathbf{z}_i, H'(\mathbf{z}_i))$ 
        \ENDIF
        \IF{$w = 1$ AND $H(\mathbf{z}_i) < \tau_c$}
           \STATE Compute loss for distribution calibration \COMMENT{Eq. 18}
            \STATE Update residuals $\boldsymbol{r}_m^c$ \COMMENT{Eq. 16-Eq. 17}
            \STATE Update Gaussian mean for inference
        \ENDIF
        \STATE Compute $p_{\text{final}}$ \COMMENT{Eq. 23}
    \ENDFOR
    \end{algorithmic}
\end{algorithm}

\section{Additional Implementation Details}

For the affine function $\mathcal{A}$ in cache-based TTA methods, we follow the settings in TDA~\cite{tda} and DPE~\cite{dpe}, applying dataset-specific hyperparameters $\alpha$ and $\beta$.
For the balance weight $\eta$ in the final inference score, we adopt the tuning strategy similar to that of TPT~\cite{tpt} and DMN~\cite{dmn}, selecting values that maximize the average validation accuracy. The optimal values typically fall within the range [0.2, 0.6]. 
The temperature parameter $\tau$ is fixed to 0.01 across all experiments.
To optimize textual residuals, we use the AdamW~\cite{adamw} optimizer with a learning rate of 0.0005, weight decay of 0.1, and $\epsilon = 10^{-3}$, applied for a single update step.


\section{Extended Ablations}

\begin{figure*}[!t]
    \centering
    \begin{tabular}{ccc}
        \includegraphics[width=0.3\textwidth]{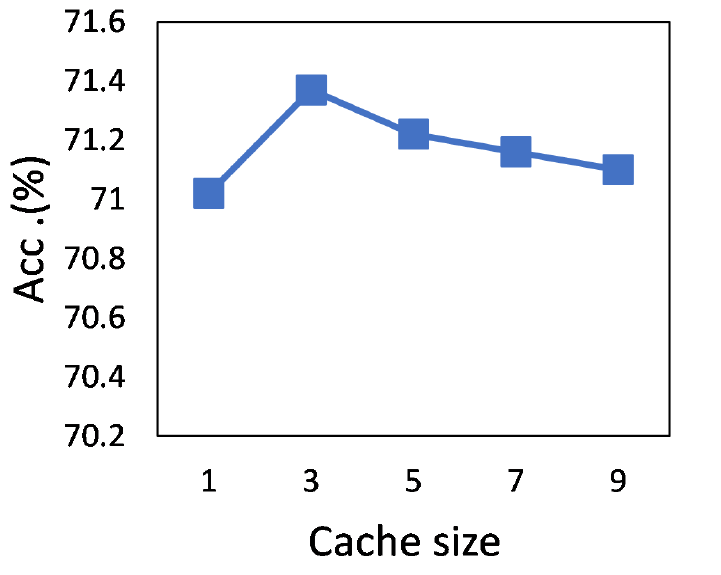}  &
        \includegraphics[width=0.3\textwidth]{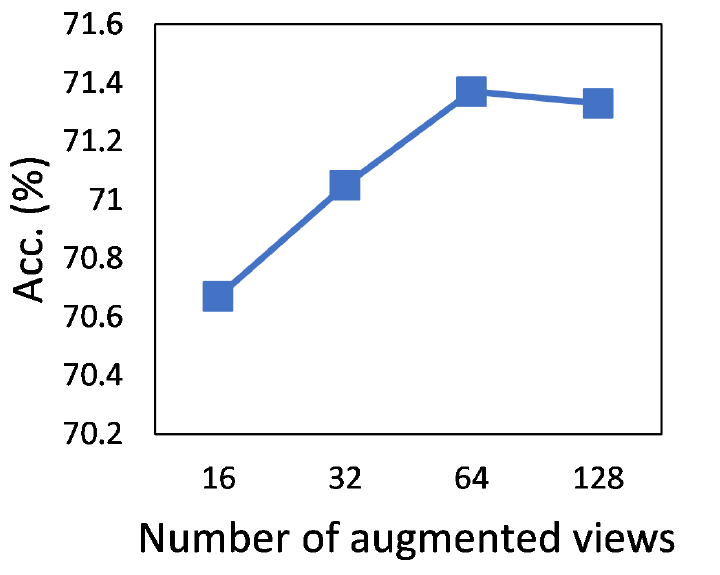}  &
        \includegraphics[width=0.3\textwidth]{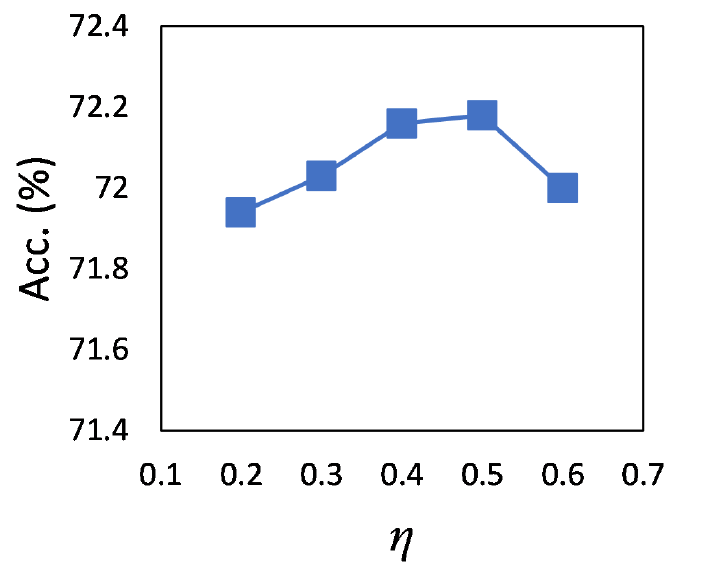}  \\
        (a)   & (b)   & (c) 
    \end{tabular}
    \caption{Sensitivity analysis of hyper-parameters. (a) Different cache size. (b) Different number of augmented views. (c) Effect of $\eta$ on ImageNet.}
    \label{fig:ablation_studies}
\end{figure*}

\textbf{Analysis of Cache Size. }Figure~\ref{fig:ablation_studies}a presents an ablation study on the effect of cache size in ReTA.  
We observe that performance improves as the cache size increases from 1 to 3, with accuracy rising from 71.02\% to 71.37\%, yielding a 0.35\% gain. However, further increasing the cache size leads to a gradual decline in accuracy, as larger caches are more likely to include lower-confidence or noisy samples.

\textbf{Ablation Studies on Augmentation View Size. }
Figure~\ref{fig:ablation_studies}b presents ReTA’s performance under varying numbers of augmentation views.
Results show performance improves with more augmentation views, but gains diminish beyond 64-128 views. 
Following TPT~\cite{tpt}, we adopt 64 views to achieve a favorable trade-off between accuracy and computational cost.

\textbf{Analysis of $\eta$. } As shown in Figure~\ref{fig:ablation_studies}c, our experiments with different values of $\eta$ (ranging from 0.2 to 0.6) show remarkably stable performance, with accuracy consistently around 72\% and a standard deviation of only 0.10\% on ImageNet.  
This low variance indicates that ReTA is not overly sensitive to the precise choice of $\eta$, enabling robust performance across a wide range of settings.  
Similar stability is observed on other datasets, suggesting that our method maintains consistent effectiveness without requiring extensive hyperparameter tuning. 

\begin{table}[!t]
    \centering
    \caption{Comparison of different formulations for the stability-consistency score.}
    \label{tab:score-formulations}
    \begin{tabular}{lc}
    \toprule
    \textbf{Formulation} & \textbf{Acc. (\%)} \\
    \midrule
    \rowcolor{gray!20}
    1 + log($\mathcal{R}\mathcal{S}$) & \textbf{71.37} \\
    $\mathcal{R} \mathcal{S}$ & 70.21 \\
    $exp{(\mathcal{R}\mathcal{S}-1)}$ & 69.45 \\
    $\sqrt{\mathcal{R}\mathcal{S}}$ & 70.95 \\
    \bottomrule
    \end{tabular}
    \label{tab:score_fun}
\end{table}

\begin{table}[!t]
    \centering
    \caption{Comparison of different formulations for the stability-consistency score.}
    \label{tab:score-formulations}
    \begin{tabular}{lc}
    \toprule
    \textbf{Formulation} & \textbf{Acc. (\%)} \\
    \midrule
    \rowcolor{gray!20}
    1 + log($\mathcal{R}\mathcal{S}$) & \textbf{71.37} \\
    $\mathcal{R} \mathcal{S}$ & 70.21 \\
    $exp{(\mathcal{R}\mathcal{S}-1)}$ & 69.45 \\
    $\sqrt{\mathcal{R}\mathcal{S}}$ & 70.95 \\
    \bottomrule
    \end{tabular}
    \label{tab:score_fun}
\end{table}

\begin{table}[!t]
    \centering
    \caption{Ablation study on the stability and consistency components.}
    \begin{tabular}{|l|c|}
    \hline
    \textbf{Parameter} & \textbf{Acc. (\%)} \\
    \hline
    \multicolumn{2}{|c|}{\textbf{Stability}} \\
    \hline
    $\gamma = 1$ & 71.15 \\
    $\gamma = 2$ & \textbf{71.37} \\
    $\gamma = 3$ & 71.25 \\
    $\gamma = 4$ & 71.22 \\
    $\gamma = 5$ & 71.14 \\
    \hline
    \multicolumn{2}{|c|}{\textbf{Consistency}} \\
    \hline
    w/ & \textbf{71.37} \\
    w/o & 70.97 \\
    \hline
    \end{tabular}
    \label{tab:cons-sta}
\end{table}

\begin{table*}[!t]
  \centering
  \caption{Comparison of Expected Calibration Error (ECE \%). Lower values indicate better model calibration.}
    \begin{tabular}{lccccccccccc}
    \toprule
    Methods & Caltech & DTD   & Cars  & EuroSAT & Aircraft & Flowers & Pets  & UCF101 & Food101 & SUN397 & Average \\
    \midrule
    TDA   & 7.45  & \textbf{35.90}  & 31.88  & \textbf{36.36}  & \textbf{35.18}  & 25.54  & 14.48  & 25.28  & 17.12  & 30.85  & 26.00 \\
    DPE   & \textbf{5.13}  & 38.51  & 30.83  & 40.65  & 44.01  & 23.04  & 13.32  & 26.75  & 15.25  & 28.73  & 26.62 \\
    \rowcolor{gray!20}
    ReTA  & 5.30  & 37.49  & \textbf{29.61} & 40.56& 39.94 & \textbf{20.85} & \textbf{9.26} & \textbf{23.18} & \textbf{13.26} & \textbf{28.33} & \textbf{24.78}\\
    \bottomrule
    \end{tabular}%
  \label{tab:ece}%
\end{table*}%

\begin{table}[!t] \small
    \centering
    \caption{Impact of different logit components on the decision boundary. \textit{Prediction Changes} (\%) indicates the proportion of samples whose predicted class changes when each logit component is removed.}
    \label{tab:decision_impact}
    \begin{tabular}{lcc}
    \toprule
    \textbf{Inference Score Composition} & \textbf{Acc. (\%)} & \textbf{Prediction Changes (\%)} \\
    \midrule
    $p_{\text{cls}} + \eta p_{\text{gauss}}$ (ReTA) & 71.37 & - \\
    ReTA - cache logits & 70.89 & 2.92 \\
    ReTA - Gauss logits & 70.51 & 6.18 \\
    \bottomrule
    \end{tabular}
\end{table}

\textbf{More Analysis About Stability-Consistency Score.}
Table \ref{tab:score_fun} compares formulations for computing the stability-consistency score. 
Our logarithmic form $1+\log(\mathcal{R}\mathcal{S})$ achieves the best accuracy (71.37\%), balancing sensitivity and robustness. 
The addition of 1 to the logarithmic term ensures that $w \geq 1$, which preserves the original entropy when the score is at its minimum. 
Other variants, such as multiplication, exponential, or square root, perform worse due to overemphasis on outliers or compressed dynamic ranges.

Table \ref{tab:cons-sta} further reports an ablation on the stability and consistency components. 
We find that $\gamma{=}2$ yields the best result, while smaller or larger values reduce accuracy due to under- or over-penalization.  
Adding consistency brings a +0.4\% gain over using stability alone, showing that both components are complementary and jointly improve performance.

\begin{table}[t!]
  \centering
\caption{Top-1 accuracy (\%) on OOD generalization. Following RLCF, we use CLIP-ViT-L/14 as the reward model and CLIP-ViT-B/16 as the base model.}
    \resizebox{\linewidth}{!}{%
    \begin{tabular}{lccccc}
    \toprule
    Method & IN-A  & IN-R  & IN-V2 & IN-Sketch & OOD Avg.  \\
    \midrule
    CLIP-ViT-B/16 & 47.87 & 73.98 & 60.86 & 46.09 & 57.20 \\
    CLIP-ViT-L/14 & 68.82 & 85.40 & 67.80 & 57.84 & 69.97 \\
    \midrule
    RLCF  & 65.45 & 83.35 & 69.77 & 54.74 & 68.33 \\
    ReTA  & 66.29 & 83.54 & 69.35 & 55.57 & 68.69 \\
    \bottomrule
    \end{tabular}%
    }
  \label{tab:rlcf}%
\end{table}%

\textbf{Comparison of Expected Calibration Error. } Table~\ref{tab:ece} reports Expected Calibration Error (ECE) computed using 20 bins, where lower values indicate better calibration. 
ReTA achieves the lowest average ECE (24.78\%), outperforming TDA (26.00\%) and DPE (26.62\%) on 7 of 10 datasets. 
With DDC refining decision boundaries via calibrated text distributions and CER improving cache reliability through consistency assessment, ReTA provides well-calibrated predictions under distribution shifts without sacrificing accuracy.

\textbf{Impact of Logit Components on Prediction and Decision Boundary. }Table \ref{tab:decision_impact} analyzes the impact of different logit components on ReTA’s decision boundary. 
Removing cache logits alters only 2.92\% of predictions, as the cache stores only a few high-confidence samples, limiting its capacity to adjust the boundary.  
In contrast, removing Gaussian logits causes 6.18\% of predictions to change and results in a larger accuracy drop, indicating that DDC plays a greater role in shaping flexible and robust decision boundaries.  
By complementing the limited coverage of cache-based methods, DDC enables more reliable predictions and significantly enhances ReTA’s adaptability under distribution shifts.

\begin{table*}[t!]
  \centering
  \caption{Top-1 accuracy (\%) results on cross-dataset generalization for EVA02-B-16. The best results are highlighted in \textbf{bold}.}
        \begin{tabular}{lccccccccccc}
        \toprule
        Method & Caltech & DTD   & Cars  & Eurosat & Aircraft & Flower & Pets  & UCF101 & Food101 & SUN397 & Average \\
        \midrule
        EVA02-B-16 & 93.02 & 52.82 & 78.55 & \textbf{66.35} & 25.06 & 72.45 & 92.11 & 68.07 & \textbf{89.43} & 70.64 & 70.85 \\
        TDA   & 96.63 & 51.48 & 79.67 & 62.56 & 26.97 & 76.25 & 92.15 & 72.32 & 86.70 & 71.49 & 71.62 \\
        DPE   & 97.20 & 59.81 & 81.22 & 66.33 & \textbf{31.83} & 76.86 & 92.67 & 73.25 & 86.81 & 73.05 & 73.90 \\
        \rowcolor{gray!20}
        ReTA  & \textbf{97.24} & \textbf{62.59} & \textbf{82.78} & 66.28 & 30.72 & \textbf{77.91} & \textbf{93.59} & \textbf{75.76} & 87.18 & \textbf{73.61} & \textbf{74.77} \\
        \bottomrule
        \end{tabular}%
  \label{tab:eva}%
\end{table*}%

\begin{table*}[t!]
  \centering
  \caption{Top-1 accuracy (\%) results on cross-dataset generalization for Coca-ViT-B-32. The best results are highlighted in \textbf{bold}.}
    \begin{tabular}{lccccccccccc}
        \toprule
        Method & Caltech & DTD   & Cars  & Eurosat & Aircraft & Flower & Pets  & UCF101 & Food101 & SUN397 & Average \\
        \midrule
    Coca-ViT-B-32 & 90.78 & 53.67 & 83.97 & 46.02 & 17.83 & 62.26 & \textbf{88.75} & 57.28 & \textbf{78.93} & 67.36 & 64.69 \\ 
    TDA   & 94.56 & 52.25 & 86.85 & \textbf{54.23} & 19.26 & 66.50  & 88.51 & 64.68 & 75.95 & 67.29 & 67.01 \\
     DPE   & 95.13 & 59.93 & 87.56 & 50.51 & \textbf{25.59} & 67.40  & 87.24 & 61.56 & 74.41 & 68.36 & 67.71 \\
     \rowcolor{gray!20}
    ReTA  & \textbf{95.33} & \textbf{61.47} & \textbf{88.97} & 50.09 & 25.32 & \textbf{69.18} & 87.38 & \textbf{67.67} & 75.69 & \textbf{68.99} & \textbf{69.01} \\
    \bottomrule
    \end{tabular}%
  \label{tab:coca}%
\end{table*}%

\begin{figure*}[!t]
\centering
\centerline{\includegraphics[width=16cm]{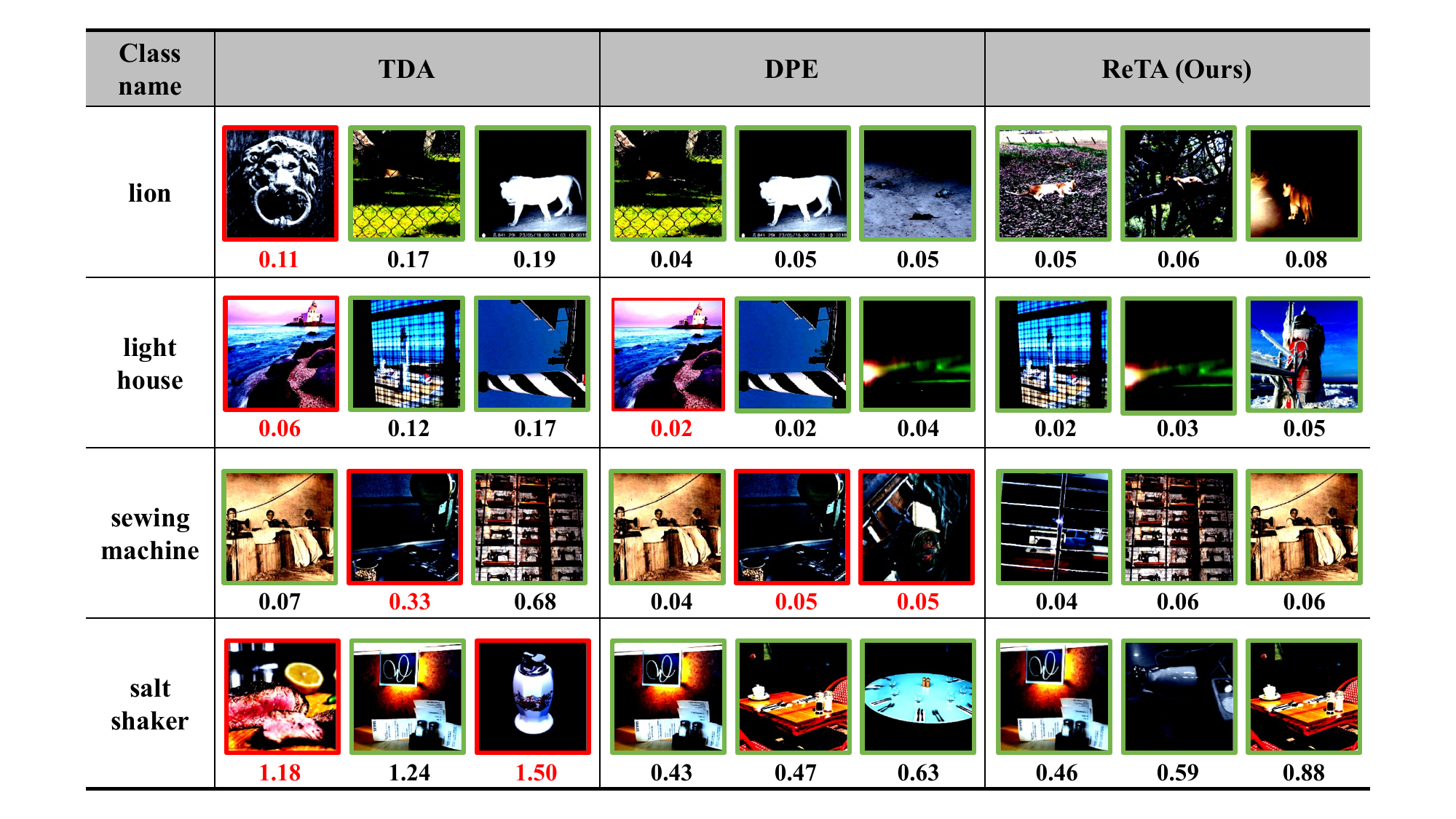}} 
    \vspace{-0.1cm}
\caption{Visual comparison of cached samples with their corresponding entropy values (shown below each image). Samples are collected after each method completes processing the ImageNet dataset. 
Green borders indicate correct classification while red borders and red entropy values denote misclassifications.}
    \label{fig:cache_visual}
\end{figure*}

\textbf{Exploring the Integration of CLIP Reward Feedback. }Motivated by the reinforcement learning with human feedback (RLHF), RLCF~\cite{rlcf} introduces a CLIP-based reward for test-time prompt tuning, using CLIP-ViT-L/14 as the reward model and CLIPScore as the reward signal. 
Following this successful approach, we modify ReTA by replacing the entropy loss with CLIPScore feedback, and use the same experimental configuration as RLCF (3 TTA steps, CLIP-ViT-B/16 as the base model).
Experiments results in Table~\ref{tab:rlcf} show that ReTA with CLIP reward achieves strong performance and surpasses RLCF by 0.36\% on average. 
This highlights not only the robustness and generalization capacity of ReTA, but also the advantage of leveraging feedback from a larger model over standard entropy-based optimization.

\textbf{Experiments on Different Models. }We conducted additional experiments using models in addition to CLIP, with the implementation from OpenCLIP~\cite{openclip} framework following~\cite{boostadapter,dpe}. We chose two alternative models: EVA-02~\cite{eva02} (with ViT-B/16) and Coca~\cite{coca} (with ViT-B/32), with the experimental results on Cross-Datasets Generalization presented in Table~\ref{tab:eva} and Table~\ref{tab:coca} respectively. 
Our proposed ReTA achieved performance improvements of 2.56\% and 1.09\% on average compared to TDA and DPE. These consistent and reliable improvements demonstrate that our approach is effective across different models, proving its effectiveness and generalizability.

\section{Examples of Hand-Crafted Prompts in Our Experiments}

\begin{figure}[!t]
\centering
\centerline{\includegraphics[width=8cm]{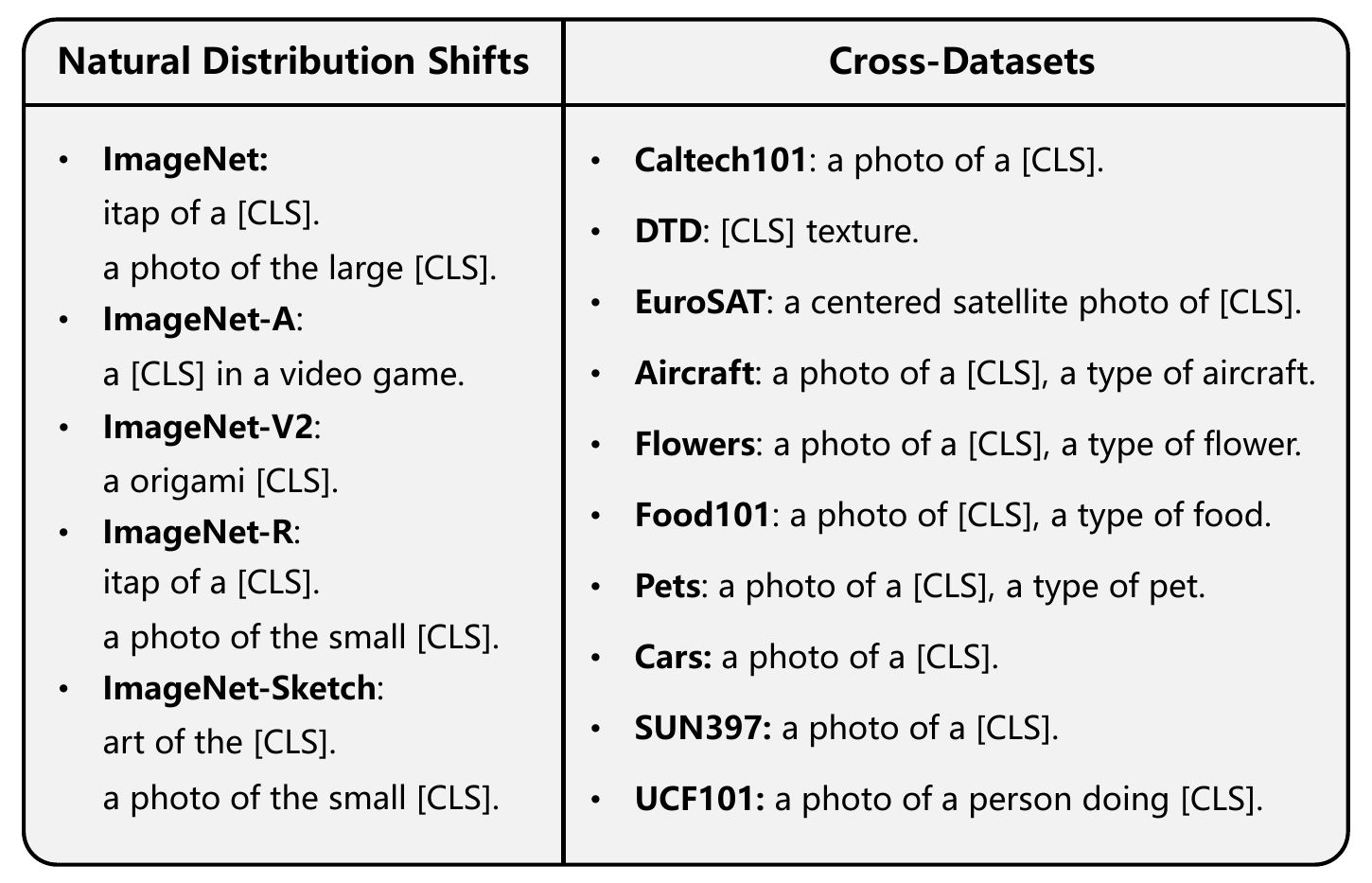}} 
    \vspace{-0.1cm}
\caption{Textual prompts examples in our experiments. In addition to these prompts, we also employ CuPL~\cite{cupl} prompts to further enhance performance.}
    \label{fig:promptexp}
    \vspace{-0.15cm}
\end{figure}

Figure~\ref{fig:promptexp} shows representative hand-crafted prompts used in our experiments.  
In addition to these, we also incorporate CuPL~\cite{cupl} prompts following DPE~\cite{dpe}.  
For each category, we ensure that the number of available prompts ($K$) exceeds the number of adjacent text embeddings ($M$), which is necessary for our ascending progressive binning strategy used to form the semantic voting committee with diverse class perspectives.

\section{Qualitative Analysis}
Figure~\ref{fig:cache_visual} presents qualitative comparisons of cached samples.
TDA often assigns low entropy to misclassified samples (\eg 0.11 for a misclassified “lion” with red border), introducing noise into the cache and leading to incorrect predictions. 
Although DPE achieves better entropy estimation through enhanced prompting, its cache still contains unreliable samples, as shown in the "sewing machine" class examples.
In contrast, ReTA consistently caches higher-quality samples with appropriate entropy values across diverse classes, validating the effectiveness of our consistency-aware entropy reweighting.

\section{More Discussion on Our Consistency-based Strategy in CER}
Self-consistency methods~\cite{consistentLLM1,consistentLLM2} have shown promising results in Large Language Models (LLMs), and tasks utilizing Vision-Language Models (VLMs)~\cite{consistencyVLM1,consistencyVLM2} have also seen improved prediction accuracy and robustness under similar guidance. 
Inspired by this success, we decided to apply a similar strategy to our method. 
Since CLIP’s predictions often show discrepancies under out-of-distribution (OOD) settings, correcting them is challenging without adjusting the backbone. Therefore, we focus on enhancing cache reliability, which is more controllable, rather than modifying the overall predictions.
We utilize consistency score as a signal for reliability, which is derived from the prediction consensus across the adjacent class-pecific text embeddings. 
Prior work~\cite{promptconsis1,promptconsis2,promptconsis3} has demonstrated the effectiveness of leveraging prompt embeddings to promote consistent predictions and improve reliability, while also showing their robustness and resilience under out-of-distribution conditions.
To ensure the validity of the consistency measurement, we introduce progressive binning approach to incorporate both similarity and diversity, ensuring that the committee formed by similar prompts offers varying perspectives. 
Additionally, we use Singular Value Decomposition (SVD) to reduce modality bias and redundancy, further enhancing the reliability of the consistency measurement. 
Through extensive experimental results, we have demonstrated that this approach, based on the consistency protocol, is both feasible and effective when applied in our CER.

\end{document}